\documentclass[10pt]{article} 
\usepackage[accepted]{tmlr}

\usepackage{amsmath,amsfonts,bm}

\def\eqref#1{equation~\ref{#1}}

\def\1{\bm{1}}

\def\mA{{\bm{A}}}

\DeclareMathAlphabet{\mathsfit}{\encodingdefault}{\sfdefault}{m}{sl}
\SetMathAlphabet{\mathsfit}{bold}{\encodingdefault}{\sfdefault}{bx}{n}

\DeclareMathOperator*{\argmin}{arg\,min}

\usepackage{latexsym}   
\usepackage{amsbsy}
\usepackage{amssymb}
\usepackage{amsmath}
\usepackage{verbatim}
\usepackage[final]{graphicx}
\usepackage{color}
\usepackage{enumitem}
\usepackage{amsthm}
\usepackage{subfigure}

\allowdisplaybreaks

\newcommand{\be}{\begin{equation}}
\newcommand{\ee}{\end{equation}}
\newcommand{\beq}{\begin{equation}}
\newcommand{\eeq}{\end{equation}}
\newcommand{\beqy}{\begin{eqnarray}}
\newcommand{\eeqy}{\end{eqnarray}}
\newcommand{\beqynn}{\begin{eqnarray*}}
\newcommand{\eeqynn}{\end{eqnarray*}}

\newcommand{\ba}{\begin{array}}
\newcommand{\ea}{\end{array}}

\newcommand{\bmx}{\begin{bmatrix}}
\newcommand{\emx}{\end{bmatrix}}
\newcommand{\bsmx}{\left[\begin{smallmatrix}}
\newcommand{\esmx}{\end{smallmatrix}\right]}
\newcommand{\bmxc}[1]{\left[\begin{array}{@{}#1@{}}}
\newcommand{\emxc}{\end{array}\right]}

\newcommand{\bt}[1]{\begin{tabular}{#1}}
\newcommand{\et}{\end{tabular}}

\newcommand{\bc}{\begin{center}}
\newcommand{\ec}{\end{center}}
\newcommand{\ben}{\begin{enumerate}}
\newcommand{\een}{\end{enumerate}}
\newcommand{\bi}{\begin{itemize}}
\newcommand{\ei}{\end{itemize}}

\newcommand{\Rbb}{{\mathbb{R}}}

\newcommand{\A}{\mathbf{A}}

\newcommand{\E}{\mathbf{E}}

\usepackage{hyperref}
\usepackage{url}
\usepackage{amssymb}            
\usepackage{mathtools}          
\usepackage{mathrsfs}           
\usepackage{graphicx}           
\usepackage{subcaption}         
\usepackage[space]{grffile}     
\usepackage{url}                
\usepackage{lipsum}             
\usepackage{algorithm}
\usepackage[noend]{algpseudocode}

\title{Incorporating Spatial Information into Goal-Conditioned Hierarchical Reinforcement Learning via Graph Representations}

\author{\name Shuyuan Zhang \email shuyuan.zhang@mail.mcgill.ca \\
      \addr School of Computer Science\\
      McGill University, Mila
      \AND
      \name Zihan Wang \email zihan.wang5@mail.mcgill.ca \\
      \addr School of Computer Science\\
      McGill University, Mila
      \AND
      \name Xiao-Wen Chang \email chang@cs.mcgill.ca\\
      \addr School of Computer Science\\
      McGill University
      \AND
      \name Doina Precup \email dprecup@cs.mcgill.ca\\
      \addr School of Computer Science\\
      McGill University, \\
      Google Deepmind,\\
      CIFAR Fellow}

\def\month{09}  
\def\year{2025} 
\def\openreview{\url{https://openreview.net/forum?id=a7Bx4s5gA8}} 

\begin{document}

\maketitle

\begin{abstract}
The integration of graphs with Goal-conditioned Hierarchical Reinforcement Learning (GCHRL) has recently gained attention, as intermediate goals (subgoals) can be effectively sampled from graphs that naturally represent the overall task structure in most RL tasks. However, existing approaches typically rely on domain-specific knowledge to construct these graphs, limiting their applicability to new tasks. Other graph-based approaches create graphs dynamically during exploration but struggle to fully utilize them, because they have problems passing the information in the graphs to newly visited states. Additionally, current GCHRL methods face challenges such as sample inefficiency and poor subgoal representation. This paper proposes a solution to these issues by developing a graph encoder-decoder to evaluate unseen states. Our proposed method, Graph-Guided sub-Goal representation Generation RL (G4RL), can be incorporated into any existing GCHRL method when operating in environments with primarily symmetric and reversible transitions to enhance performance across this class of problems. We show that the graph encoder-decoder can be effectively implemented using a network trained on the state graph generated during exploration. Empirical results indicate that leveraging high and low-level intrinsic rewards from the graph encoder-decoder significantly enhances the performance of state-of-the-art GCHRL approaches with an extra small computational cost in dense and sparse reward environments.
\end{abstract}

\section{Introduction}
\label{sec:introduction}
Traditional reinforcement learning methods face great challenges when learning policies in environments with long time horizons and sparse rewards. 
To address these challenges, Hierarchical Reinforcement Learning (HRL) methods have been proposed to break problems into smaller, more manageable subproblems conducive to more efficient learning. 
Previous works \citep{sutton1999between,kulkarni2016hierarchical,vezhnevets2017feudal, levy2017learning} have demonstrated HRL’s capability in handling large, sparse-reward environments. 
Among HRL methods, Goal-Conditioned Hierarchical Reinforcement Learning (GCHRL) has attracted much attention due to its well-defined paradigm \citep{nachum2018data} and its resemblance to the human thinking process \citep{zhao2022simulate, hsu2024thought}. Although GCHRL has shown superior performance compared to non-hierarchical methods in some scenarios, questions such as how to learn better subgoal representations and explore the state space more efficiently \citep{nachum2018near, guo2021state} remain unanswered.

GCHRL methods typically utilize two levels of agents as described in \citet{nachum2018data}.
The high-level agent chooses the next target state based on its current state,
while the low-level agent decides how to reach this target. 
Both agents face their own challenges. 
The high-level agent suffers from sample inefficiency in environments with a large state-action space, a problem also encountered by non-hierarchical methods,
while the low-level agent is trained solely by the reward signals derived from the distances between the current state and subgoals in the representation space, making the reward signals highly susceptible to poor subgoal representations.

Previous works have tried to either enhance the efficiency of high-level exploration \citep{huang2019mapping,zhang2022adjacency} or find a good subgoal space to boost GCHRL's performance in complex environments \citep{wang2024probabilistic}. To the best of our knowledge, while these methods have made progress in their respective aspects, no prior work has attempted to address the efficiency of high-level exploration and the accuracy of low-level learning signals in a unified framework.
Yet, such integration is crucial, as increasing the effectiveness of the high-level agent and the accuracy of the low-level agent within a cohesive framework can yield performance improvements that surpass simple additive effects.

Recently, combining graph theory with RL has become a trend in the community \citep{lee2022dhrl, gieselmann2021planning} as graphs are inherently well-suited for representing the environment and task structure. 

Previous work has focused on areas such as decision-making through graph search or traversal \citep{wan2021reasoning, shang2019learning, eysenbach2019search}, and as well as using graphs as world models \citep{zhang2021world, huang2019mapping}.
Yet, many previous works rely on pre-crafted graphs, limiting their generalizability. Additionally, most existing works \citep{zhu2022value, hong2022topological} constructed graphs directly from the original state space,
which cannot provide meaningful guidance when the agent encounters a state that is not represented in those graphs.

To utilize graph representations when a new state (node)  is encountered, transitioning to these representations \citep{hamilton2020graph, chen2020graph, khoshraftar2024survey} through graph learning is a viable option. 
Recent studies have shown that employing graph representations 
for learning can improve underlying RL performance \citep{klissarov2020reward,klissarov2023deep}.

In this paper, we propose a novel approach that simultaneously addresses all of the aforementioned problems.
Specifically,  
we construct a state graph based on visited states until the number of nodes reaches a threshold and after that we update the graph by adding and dropping nodes as new states are visited.
Using this graph, we build a subgoal space through graph learning that leverages both state representation and spatial connectivity. 
By generating subgoal representations through graph learning,
we ensure that they capture their relative positions 
in the decision chain, thereby forming a more effective subgoal space. 
In estimating the distance between the current state and the intended subgoal in the original space, we use the distance between their corresponding representations in the subgoal space. 
This distance is then used to calculate the intrinsic reward for curiosity-driven exploration, aiming to improve sample efficiency across both high-level and low-level agents. 

The main contributions of this paper are as follows:
\begin{itemize}
  \item We propose a novel architecture that employs a graph encoder-decoder to embed spatial information into subgoal representations, enabling the evaluation of newly visited states. This architecture can be integrated into any GCHRL algorithm operating in environments with primarily symmetric and reversible transitions, enhancing performance across this class of problems.
  
  \item We present a method for the online construction of the state graph 
  as a graph-based world model \citep{ha2018world,zhang2021world} for the HRL agent by sampling from trajectories.
  
  \item We use novelty-based auxiliary rewards \citep{csimcsek2006intrinsic, nehmzow2013novelty} derived from subgoal representations to improve sample efficiency for both high-level and low-level agents.
\end{itemize}
We tested our approach on several environments \citep{todorov2012mujoco} to assess the significance of our experimental results. The findings indicate that our method can significantly enhance 
the performance of the underlying HRL approach in terms of both sample efficiency and success rate/cumulative reward.

\section{Preliminaries}
\label{sec:pre}
\subsection{Markov Decision Processes}
As the most common framework for modeling reinforcement learning scenarios, the Markov Decision Process (MDP) \citep{puterman2014markov} is introduced as a tuple $<\mathcal{S}, \mathcal{A}, P, R, \gamma>$, defined as follows: At each time step $t$,  the agent observes the current state $s_t \in \mathcal{S}$ provided by the environment and chooses an action $a_t \in \mathcal{A}$ according to its internal policy $\pi(a_t|s_t)$, which specifies the probability of choosing action $a_t$ given state $s_t$. 
The action is then executed, and the interaction with the environment leads the agent to a new state $s_{t+1}$ according to a transition probability function $P(s_{t+1}|s_t, a_t)$ which is known only to the environment.  
Subsequently, the agent receives a reward $r_t$, determined by the reward function $R(s_t,a_t)$ that evaluates the action taken in the current state and is also only visible to the environment. The agent aims to learn an optimal policy $\pi$ to maximize the expected discounted cumulative reward $\mathbb{E}_{\pi} \big[\sum_{t=0}^T \gamma^t r_t \big] $, where $\gamma$  (with $0\leq\gamma<1$) is a pre-defined discount factor used to prioritize immediate rewards over distant future rewards, thereby ensuring that the total reward remains finite.

\subsection{Goal-conditioned Hierarchical RL (GCHRL)}

Goal-Conditioned Reinforcement Learning (GCRL) trains agents to achieve specific goals, which are the target states. The agent receives an additional goal input $g_t$ along with the state input $s_t$ and learns a policy $\pi(a_t|g_t, s_t)$ that aims to achieve this goal. Goals are represented explicitly in the input to the policy, guiding the agent’s actions towards desired outcomes.
The reward function is often goal-dependent, providing positive feedback when the agent successfully reaches the desired goal state.



To deal with large and complex environments, 
Goal-Conditioned Hierarchical Reinforcement Learning (GCHRL) \citep{nachum2018data, zhang2022adjacency, wang2024probabilistic} decomposes the learning task into a hierarchy of smaller, more manageable sub-tasks. Typically, there are two levels of agents. At time step $t$, the high-level agent chooses a subgoal $g_t$, a representation of a target state, 
and assigns it to the low-level agent to achieve as part of the overall task. 
This choice is made by sampling $g_t$ from the high-level policy $\pi_h(g_t|\phi(s_t))$,  where $\phi: s \mapsto \mathbb{R}^d$ is the state representation function which gives a condensed representation of the state. 


Each state $s_t$ can be mapped to its subgoal feature $g(s_t)$ by a subgoal feature extractor. 
Note that $g(s_t)$ is not the same as $g_t$. 
The former, $g(s_t)$, is the learned subgoal feature of the current state $s_t$, while the latter, $g_t$, is the target state we aim to reach from the state $s_t$ in one step or multiple steps.

Given the subgoal $g_t$ sampled from 
the high-level policy  $\pi_h(g_t|\phi(s_t))$ for the current time step $t$ and the state representation vector $\phi(s_t)$, a low-level agent executes action $a_t$ based on the low-level policy $\pi_l(a_t|\phi(s_t), g_t)$. The low-level agent is trained using the intrinsic reward signal $r_{\text{int}}(s_t, g_t, a_t, s_{t+1})=-\|\phi(s_{t+1})-g_t\|^2_2$ to encourage it to achieve the subgoal.

Both agents can be implemented by any policy-based methods, including those introduced in previous works on policy gradients such as \cite{fujimoto2018addressing, haarnoja2018soft} and \cite{schulman2017proximal}.

\subsection{Graph and MDP}
Graph is a generic data structure, which can model complex relations among objects in many real-world problems. 
A graph is defined as $\mathcal{G}=(\mathcal{V},\mathcal{E})$, where $\mathcal{V}=\{1,2,\ldots,N\}$ is the set of nodes and $\mathcal{E}=\{e_{ij}\}$ is the set of edges without self-loops. The adjacency matrix of $\mathcal{G}$ is denoted by $\mA=(\mA_{i,j})\in \Rbb^{N\times N}$ with $\mA_{i,j}=1$ if there is an edge between nodes $i$ and $j$, otherwise $\mA_{i,j}=0$. 
The adjacency matrix can be extended to a {\em weighted} adjacency matrix, 
where $\mA_{i,j}$ is a weight of the edge $e_{ij}$. 

In MDP, a node can represent a state, while the edge weights can model the transition probabilities or reachability statistics between states.

\section{Methods}
\label{sec:methods}

This section presents our framework, Graph-Guided subGoal representation Generation (G4RL). 
Our method reshapes the subgoal space utilizing a state graph to incorporate the relative spatial information of visited states.

One drawback of previous hierarchical reinforcement learning algorithms \citep{nachum2018data,kim2021landmark,zhang2022adjacency,luo2024goal} 
is that the Euclidean distance  
calculated in the original state representation space between the current state and the intended goal does not accurately reflect the true progress of the low-level agent, as there is rarely a straight path between the current state and the subgoal in the space. 
As a result, the low-level agent trained with such information may receive an inaccurate reward signal, thus impairing its performance. 
Another issue is that, without appropriate constraints,
the high-level agent may propose subgoals that are too difficult to reach, 
wasting exploration steps on pursuing infeasible targets \citep{zhang2022adjacency}. 
Our proposed method aims to mitigate both problems by calculating the distance in a subgoal representation space between subgoal representations given by a graph encoder-decoder. 
This graph encoder-decoder captures the actual connectivity between states, ensuring that the generated subgoal representations respect adjacency information.

\subsection{State graph}
To record the visited states and their connections, we maintain a state graph $\mathcal{G}=(\mathcal{V},\mathcal{E})$ with a fixed number $N$ of nodes\footnote{The number of training 
states for the graph encoder-decoder grows quadratically with $N$ because the adjacency weight matrix has $N^2$ elements. The choice of $N$ depends on the machine's capabilities.}. This graph is built and updated during training, with no pre-training using expert data or handcrafted process involved in its construction.

Each node is labelled by the corresponding state and for each node $s_t$, the corresponding state representation vector $\phi(s_t)$, which is also referred to as the node feature, 
is stored.  Edges in the graph represent connectivity between states. The graph is constantly updated during exploration.

We choose the state graph to be undirected for mainly two reasons: \textbf{(1) Efficiency:} An undirected graph requires fewer resources for storage and computation compared to a directed graph, and \textbf{(2) Method compatibility:} Our method relies on defining a deterministic distance between state/subgoal representations for each node pair. This is straightforward in an undirected graph but becomes problematic in a directed setting, where distances can be asymmetric or undefined. Related details will be explained in Sections \ref{sec:GED} and \ref{sec:AT}.

This choice is based on the assumption that G4RL is designed for environments with symmetrical and reversible dynamics. However, some of our experiments demonstrate that G4RL can still enhance performance in partially asymmetric environments; please refer to the experiments section for further details.

\subsubsection{Graph construction}\label{sec:construction}
The graph is initialized with $N$ empty nodes and no edges. The corresponding weighted adjacency matrix $\mA$ is set to an $N\times N$ zero matrix. 
We perform the GCHRL exploration process using randomly initialized policy $\pi_h$ and $\pi_l$. Once the agent encounters a state representation never seen before, that is, the representation is different from any state representations stored in the graph,
as described in equation (\ref{eq:newnode}), it stores the state representation $\phi(s_t)$ as the node feature of an empty node
in the graph and build an edge between this node and the node corresponds to the previous state:
\begin{equation} \label{eq:newnode}
\forall_{s_v \in \mathcal{V}},   \|\phi(s_t)-\phi(s_v)\|_2>\epsilon_d,
\end{equation}
\begin{equation} \label{eq:newnodeedge}
\mA_{\phi(s_t), \phi(s_{t-1})}=\mA_{\phi(s_{t-1}), \phi(s_t)}=1,
\end{equation}
where 
$\epsilon_d$ is a hyper-parameter controlling the distance threshold between state representations.
When the agent encounters a state $s_t$ with feature $\phi(s_t)$ that is similar to several node representations already stored in the graph, it finds the state whose representation is
the closest to the current state feature:
\begin{equation} \label{eq:oldnode}
s_v = \argmin_{s_u: \|\phi(s_t)-\phi(s_u)\|_2 \leq \epsilon_d} \|\phi(s_t)-\phi(s_u)\|_2.
\end{equation}
Then the node $s_v$ is relabeled as $s_t$ and the weight for the edge $(s_{t-1},s_t)$
is updated as follows:
\begin{equation} \label{eq:updateedge}
\mA_{\phi(s_{t-1}), \phi(s_t)}=\mA_{\phi(s_t), \phi(s_{t-1})}:=\mA_{\phi(s_{t-1}),\phi(s_t)}+1.
\end{equation}
Note that a large weight indicates more frequent transitions between the underlying states.

We have used the Euclidean norm to define the distance between feature vectors.  Since some elements may contain more spatial information than others, 
one can use a weighted Euclidean norm to define the distance between state representations instead. 

\subsubsection{Graph updating}
The graph has a fixed number of nodes.
Suppose the graph is now full.
When a new state $s_t$ is encountered, 
if $s_v$ from equation (\ref{eq:oldnode}) exists, as before we relabel the node as $s_t$ 
and perform edge update as shown in equation (\ref{eq:updateedge});
Otherwise, we replace the oldest state node in the graph with the current state node,
delete all edges previously linked to that node,
and create an edge $(\phi(s_{t-1}),\phi(s_t))$ with weight $\A_{\phi(s_{t-1}),\phi(s_t)}=\A_{\phi(s_t),\phi(s_{t-1})}=1$.
Alternatively, we could replace the state node that is most weakly connected to the other nodes--that is, the node with the lowest sum of edge weights.

\subsection{Graph encoder-decoder}
\label{sec:GED}
To enable the assignment of suitable subgoal representations to every possible state, including unseen ones, we use node representations and edges to train a graph encoder-decoder. The parameter updates of the graph encoder-decoder and the policies during policy training
are performed alternately in each episode. 

The encoder-decoder starts training after the graph is full and continues periodically after processing a few trajectories. Section 3.3 will show the details of the training schedule.

The encoder $\E$  maps every state representation $\phi(s)$ to a subgoal representation $g(s)$. 
We use a feed-forward network (FFN) 
with several layers as the encoder $\mathbf{E}$:
\begin{equation} \label{eq:graphe}
g(s)=\mathbf{E}(\phi(s))=\text{FFN}(\phi(s)).
\end{equation}
The weight parameters of the feed-forward network will be learned through training.
The decoder $\mathbf{D}$ takes two subgoal representations as input
and outputs the inner product of these two representations:
\begin{equation} \label{eq:graphd}
\mathbf{D}(g(s_u),g(s_v)) = g(s_u)^Tg(s_v).
\end{equation}
We choose dot-product similarity based on the assumption that the similarity between two nodes, such as the overlap in their local neighbourhoods, is well captured by the dot product of their embeddings. This assumption is supported by prior work in the graph embedding literature \citep{ahmed2013distributed, cao2015grarep, ou2016asymmetric}.

The aim is to use the encoder-decoder structure to predict node relations. 
Naturally we can use $\mA_{\phi(s_u), \phi(s_v)}$ as a measure of the relation 
between nodes $\phi(s_u)$ and $\phi(s_v)$. 
But for the sake of numerical stability in the training process,
we use 
$\mA_{\phi(s_u),\phi(s_v)}/\max_{\phi(s_i),\phi(s_j)}\mA_{\phi(s_i),\phi(s_j)}$
as a measure. 
Thus the loss function is defined as:
\begin{equation} \label{eq:edloss}
    \mathcal{L}=\sum_{\phi(s_u),\phi(s_v) \in \mathcal{V}} 
    \big[ \mathbf{D}(g(s_u),g(s_v))-\mA_{\phi(s_u), \phi(s_v)}/\max_{\phi(s_i),\phi(s_j)}\mA_{\phi(s_i),\phi(s_j)} \big] ^2.
\end{equation}
This loss function can enforce the subgoal representation provided by the encoder
to respect neighbouring features in the graph. 

Note that in each training phase of the graph encoder-decoder (except the first one), we use the values of the parameters obtained from the last training phase as the initial point, which helps save computation cost.

\subsection{Adaptive training schedule of the graph encoder-decoder}
\label{sec:AT}
The graph stores evolving data, including state representations as node features and connection information in the weighted adjacency matrix $\mA$, which are continuously updated during online training. Since the graph structure and content change at varying rates across episodes, training the graph encoder-decoder at fixed intervals can cause several issues:
(1) high variance in earlier episodes, where sparse or unstable graph data may lead to unreliable model updates;
(2) data underutilization, where intermediate graph states are overwritten before being used for training; and
(3) overfitting in later episodes, as the model repeatedly trains on increasingly redundant data.
To address these issues, we introduce an adaptive training schedule for the graph encoder-decoder, described in the following paragraph.

There are two types of data changing in the graph: node replacement
and edge update. 
We introduce a variable $c$ to track the weighted number of data changes. 
Since the replacement of nodes has a much higher impact on the data than the edge update, we add $N-1$ to $c$ if a node replacement occurs, and add $1$ to $c$ if an edge update happens:
\begin{equation} \label{eq:updateschedule}
  c =
    \begin{cases}
      c + N-1, & \text{if a node replacement happens,}\\
      c + 1, & \text{if an edge update happens.} 
    \end{cases}       
\end{equation}
When this variable exceeds a certain value, specifically a tolerance $\beta$ multiplied by
the total number of non-diagonal elements $N^2-N$ in the matrix $\mA$. 
we perform one gradient update for the graph encoder-decoder,  and
then we reset $c$ to $0$.

\subsection{Hierarchical agent with graph encoder-decoder}
Our proposed method involves traditional goal-conditioned settings and a subgoal representation extractor implemented by a graph encoder-decoder.

The high-level policy $\pi_h(g_t|\phi(s_t))$
nominates a subgoal every $K$ steps and is trained using the external environmental reward $r_{\text{ext}}$. 

The policy can be implemented by any policy-based RL algorithm that takes transition tuples $(s_t, g_t, a_t, r_t, s_{t+1}, g_{t+1})$ as input. 
To encourage it to propose a subgoal that is not too difficult to reach from the current state $s_t$ for more efficient exploration, 
we add an intrinsic term to the high-level reward, considering the distance between the subgoal features of $s_t$ and $g_t$ in the subgoal space:
\begin{equation} \label{eq:hir}
r_h(s_t, g_t, s_{t+1})= r_{\text{ext}}+r_{\text{int}}=r_{\text{ext}}+\alpha_h \cdot \mathbf{D}(\mathbf{E}(\phi(s_t)), \mathbf{E}(g_t)),
\end{equation}
where $\alpha_h$ is a hyperparameter that controls the significance
of the intrinsic term in the high-level reward.

The low-level policy $\pi_l(a_t|\phi(s_t), g_t)$, 
however, operates in the subgoal space. While it still takes $\phi(s_t)$ and $g_t$ as input and outputs an atomic action $a_t$, we compute the reward based on distances in the subgoal space:
\begin{equation} \label{eq:lir}
r_l(s_t, g_t, a_t, s_{t+1})=-\|\phi(s_{t+1})-g_t\|^2 + \alpha_l \cdot \mathbf{D}(\mathbf{E}(\phi(s_{t+1})), \mathbf{E}(g_t)),
\end{equation}
where $\alpha_l$ is a hyperparameter controlling the significance of the reward term in the low-level reward. By computing the intrinsic reward in the subgoal space rather than in the state space, the function provides high values when proposed subgoals are easy to reach from the current location and low values when subgoals are close in the original state space but difficult to reach from the current location. The low-level agent can also be any policy-based algorithm.

\subsection{Balancing between speed and performance}\label{ss:balance}
Due to the excessive comparisons between the current state representation and the node features during graph updates,
as well as the training cost of the graph encoder-decoder on a large graph, our experiments show that the GCHRL method, after incorporating our method, takes approximately twice as long as the original GCHRL method.

To reduce the additional cost, we can 
either decrease the frequency of sampling candidates for node features, train the graph encoder-decoder with a subset of all available training data, or do both.

For the sampling frequency, instead of comparing the state representation with node features 
in each time step, we do it in every $t_c$ time steps.
This may significantly speed up our method while maintaining  satisfactory performance.

To reduce the training data in each graph encoder-decoder training cycle, we randomly sample node pairs in the graph instead of using every node pair for training.

In the experiments section, we will present the training time and performance of G4RL with the above two techniques applied.

\subsection{Algorithm: GCHRL + G4RL}

The detailed description of how our proposed strategy, G4RL, can be incorporated into typical GCHRL algorithms is given in  Appendix \ref{sec:algo}.



\section{Experiments} \label{s:exp}
In this section, we empirically evaluate the effectiveness of integrating G4RL into existing GCHRL methods. The experiment results demonstrate substantial improvements in both convergence speed and overall success rates achieved by our proposed approach. Additionally, we provide empirical evidence that the generated state graph accurately represents the underlying structure and relationships within the task environments.

\subsection{Environment settings}
We used AntMaze, AntGather, AntPush, AntFall and AntMaze-Sparse environments from the GYM MuJoCo library \citep{todorov2012mujoco}. The first four involve complex navigation and manipulation tasks performed by a simulated multi-armed robot, while AntMaze-Sparse presents a particularly challenging scenario due to sparse reward signals, providing feedback only upon reaching the goal. Note that AntPush and AntFall contain asymmetric transitions which lead to irreversible state changes that can be difficult to learn with an undirected graph. We added these environments to show G4RL's robustness in inherently asymmetric environments. For the state representation, we selected a subset of raw state dimensions that contains spatial information (e.g. coordinates and arm angles) to serve as the node representation for the graph encoder-decoder. We deliberately aligned our choice of environments with prior work to ensure a fair and consistent comparison between the backbone algorithms and their G4RL-augmented versions. This allowed us to use the same environments, hyperparameters, and codebases provided by the original studies. Our goal was to demonstrate that G4RL can consistently enhance the performance of these backbone algorithms under comparable conditions.

\subsection{Experimental Comparisons}
We incorporated G4RL in the following existing GCHRL methods:
\begin{itemize}
    \item \textbf{HIRO} \citep{nachum2018data}: This is the first method which describes how the Goal-conditioned information can be integrated into hierarchical agents.
    \item \textbf{HRAC} \citep{zhang2022adjacency}: This method enhances the performance of HIRO by training an adjacency network that produces subgoals easier to reach from the current subgoal. 
    \item \textbf{HESS} \citep{li2022active}: This method applies a regularization term on consecutive subgoal representations in each update to stabilize the representation across episodes. 
    \item \textbf{HLPS} \citep{wang2024probabilistic}: This method applies the Gaussian process on subgoal representations for a smoother representation update.
\end{itemize}
In addition to comparing these four GCHRL-G4RL methods with their native counterparts,  we also compared them with the following non-hierarchical method:
\begin{itemize}
    \item \textbf{TD3} \citep{fujimoto2018addressing}: This is a well-known non-hierarchical policy-based method designed for continuous action spaces and we use it to implement both high- and low-level agents. 
\end{itemize}

Although reward is a key metric of an agent's learning ability,
for AntMaze and AntMaze-Sparse, we compare success rates of  these methods 
instead of their rewards on the corresponding tasks.
This is because higher rewards in AntMaze/AntMaze-Sparse do not necessarily indicate better performance; the agent may achieve high rewards without reaching the goal.

\begin{figure}[h!]
    \centering
    \subfigure[]{\includegraphics[width=0.32\textwidth]{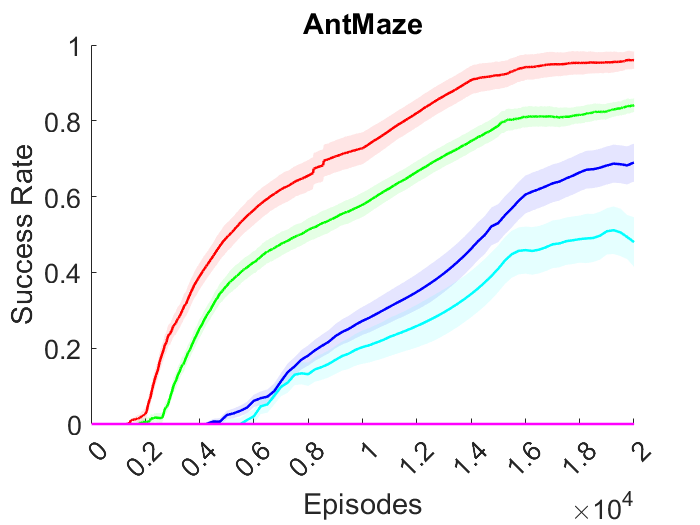}} 
    \subfigure[]{\includegraphics[width=0.32\textwidth]{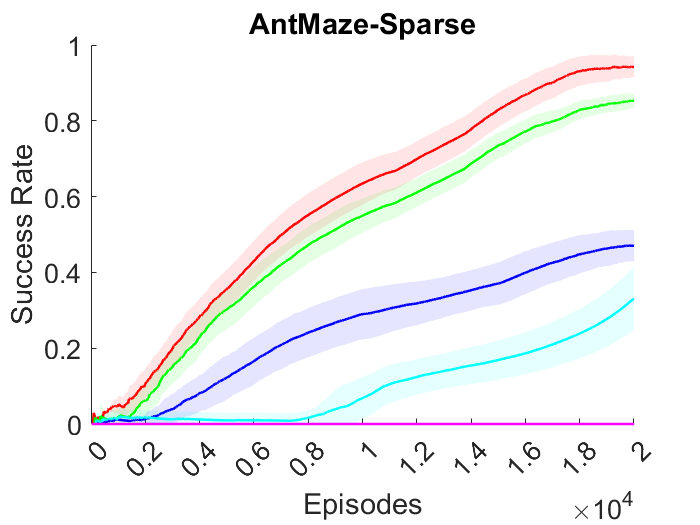}} 
    \subfigure[]{\includegraphics[width=0.32\textwidth]{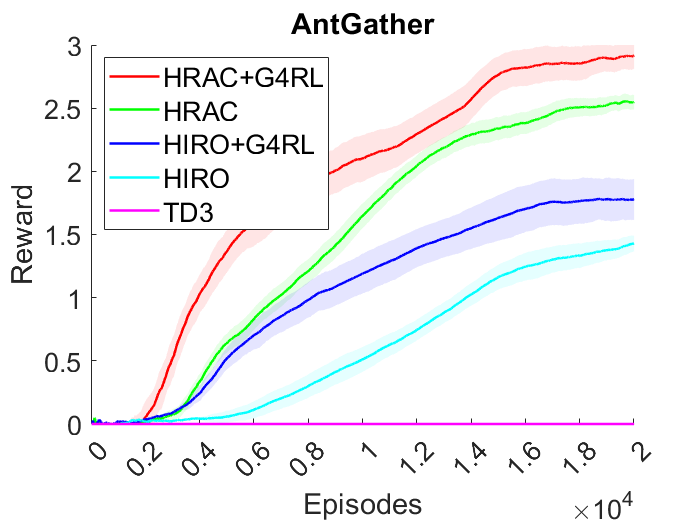}}
    \caption{Success Rate on (a) AntMaze (b) AntMaze-Sparse 
    and Reward on (c) AntGather, using HIRO, HIRO-G4RL, HRAC, HRAC-G4RL, and TD3. Incorporating G4RL in HIRO and HRAC significantly enhances their performance.}
    \label{fig:comp}
\end{figure}

\begin{figure}[h!]
    \centering
    \subfigure[]{\includegraphics[width=0.32\textwidth]{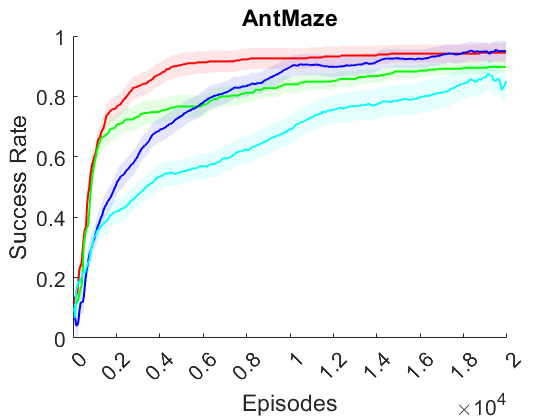}} 
    \subfigure[]{\includegraphics[width=0.32\textwidth]{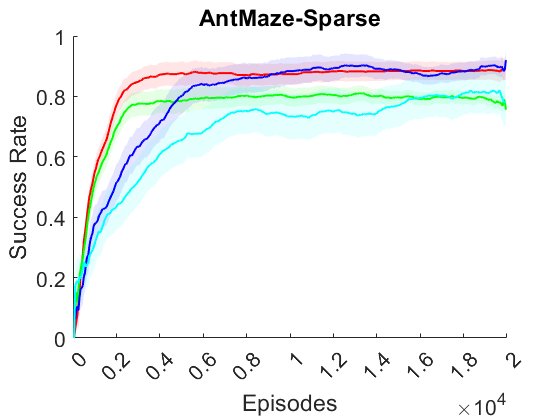}} 
    \subfigure[]{\includegraphics[width=0.32\textwidth]{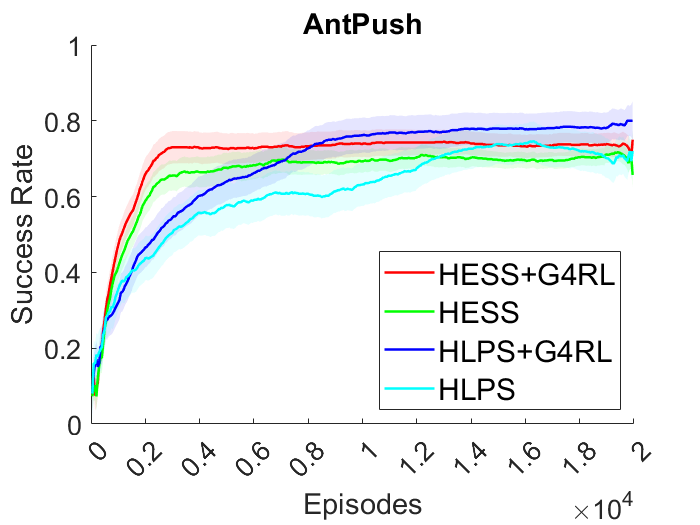}}
    \caption{Success Rate on (a) AntMaze (b) AntMaze-Sparse 
    and (c) AntPush, using HESS, HESS-G4RL, HLPS, HLPS-G4RL. Incorporating G4RL in HESS and HLPS significantly enhances their performance.}
    \label{fig:comp1}
\end{figure}

The learning curves of baseline methods and G4RL-applied versions are plotted in Figure \ref{fig:comp} and \ref{fig:comp1}. Note that all the curves reported in Section \ref{s:exp} are averages from 20 independent runs and they have been equally smoothed for better visualization.

From Figures \ref{fig:comp} and \ref{fig:comp1}, we observe that, in all environments,
incorporating G4RL in the base GCHRL methods significantly enhances their performance, further improving the already strong results of these hierarchical methods 
compared to the non-hierarchical method. 
Notably, G4RL-augmented methods not only achieve higher final success rates but also converge substantially faster, with the most significant improvements observed during the early stages of training.

\begin{figure}[h!]
    \centering
    \subfigure[]{\includegraphics[width=0.32\textwidth]{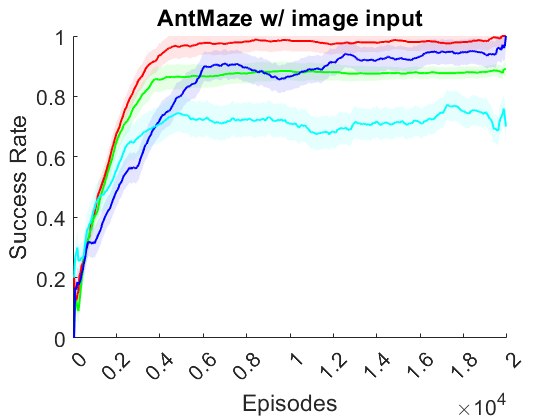}} 
    \subfigure[]{\includegraphics[width=0.32\textwidth]{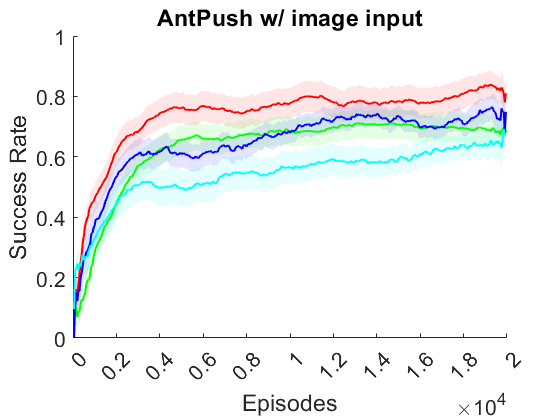}} 
    \subfigure[]{\includegraphics[width=0.32\textwidth]{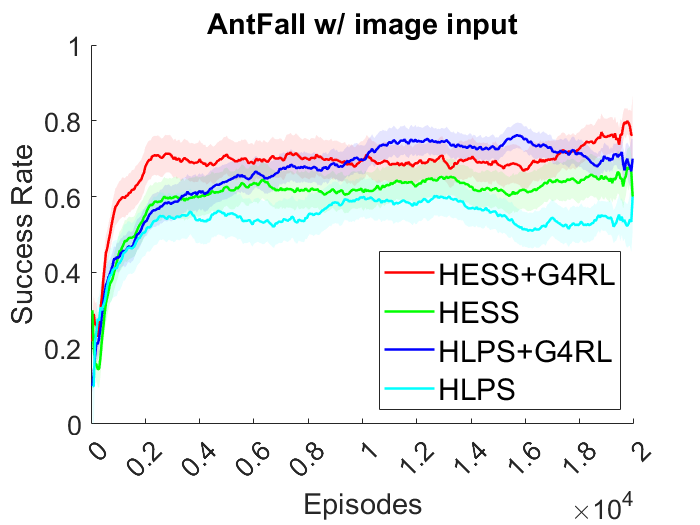}}
    \caption{Success Rate on (a) AntMaze (b) AntPush 
    and (c) AntFall with image state features, using HESS, HESS-G4RL, HLPS, HLPS-G4RL. Incorporating G4RL helps convergence and achieves higher performance across all tested image-based environments.}
    \label{fig:image}
\end{figure}

To demonstrate the proposed method's effectiveness in environments with image-based state representations, we conducted experiments on AntMaze, AntPush, and AntFall, utilizing images as states, and compared the results with HESS and HLPS, along with their G4RL variations. We use Mean Squared Error (MSE) to measure the pixel-wise differences between image states to decide whether a new node should be added to the graph. 
The test results, given in Figure \ref{fig:image}, show that methods incorporating G4RL exhibit faster convergence and achieve higher performance across all tested image-based environments.

\subsection{Ablation study}
\subsubsection{The effect of high/low-level intrinsic reward}
We consider the following variants of G4RL to show the effectiveness of adding high-level and low-level intrinsic rewards:
\begin{itemize}
    \item \textbf{High+Low-level intrinsics}: Apply both equation (\ref{eq:hir}) and equation (\ref{eq:lir}) to the high-level and low-level rewards respectively.
    \item \textbf{High-level intrinsic only}: Apply equation (\ref{eq:hir}) to the high-level rewards and set $\alpha_l=0$ in equation (\ref{eq:lir}) when it is applied 
    to the low-level rewards. 
    \item \textbf{Low-level intrinsic only}: Apply equation (\ref{eq:lir}) to the low-level rewards and 
    set $\alpha_h=0$ in equation (\ref{eq:hir}) when it is applied  to the high-level rewards.
    \item \textbf{HIRO/HRAC/HESS/HLPS}:  Vanilla baseline methods.
\end{itemize}

Same as before, all the curves reported in this section are drawn from results averaged across 20 independent runs. All curves have been equally smoothed for better visualization. Additionally, we have added the un-smoothed versions of figures \ref{fig:HIRO_ab} and \ref{fig:HRAC_ab} in Appendix \ref{sec:unsmooth} to confirm that the conclusions are unaffected by the smoothing.

\begin{figure}[h!]
    \centering
    \subfigure[]{\includegraphics[width=0.32\textwidth]{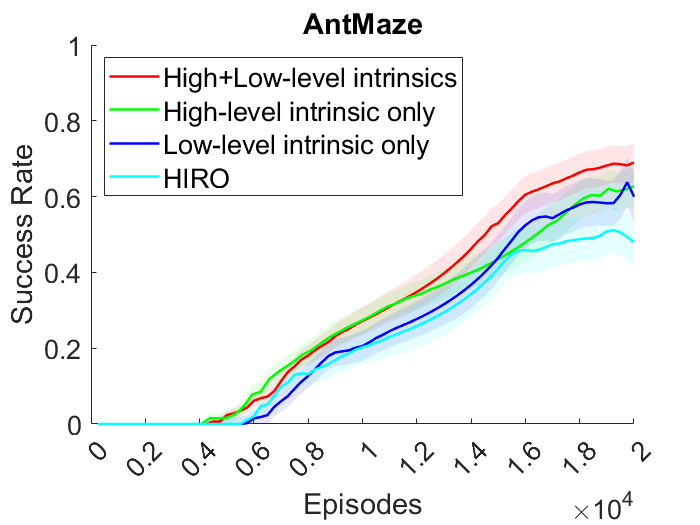}} 
    \subfigure[]{\includegraphics[width=0.32\textwidth]{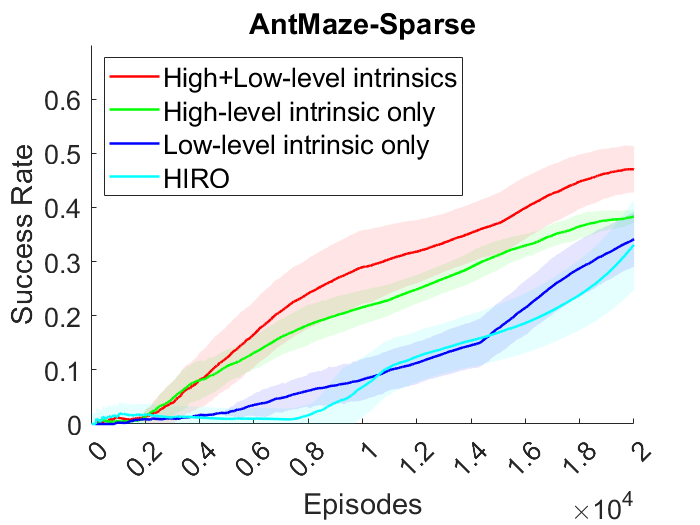}} 
    \subfigure[]{\includegraphics[width=0.32\textwidth]{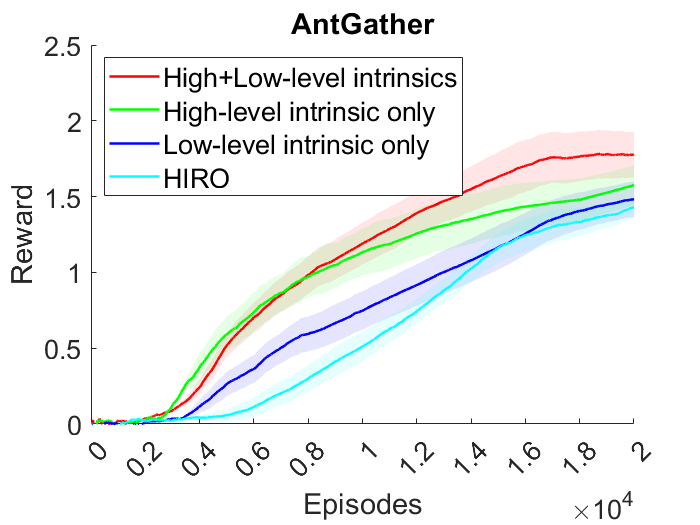}}
    \caption{Success Rate on (a) AntMaze (b) AntMaze-Sparse 
    and Reward on (c) AntGather using HIRO-G4RL, HIRO + High-level intrinsic, HIRO + Low-level intrinsic and HIRO. All curves have been equally smoothed for better visualization. The combination of high-level and low-level intrinsic rewards results in the highest success rates and fastest convergence.}
    \label{fig:HIRO_ab}
\end{figure}

\begin{figure}[h!]
    \centering
    \subfigure[]{\includegraphics[width=0.32\textwidth]{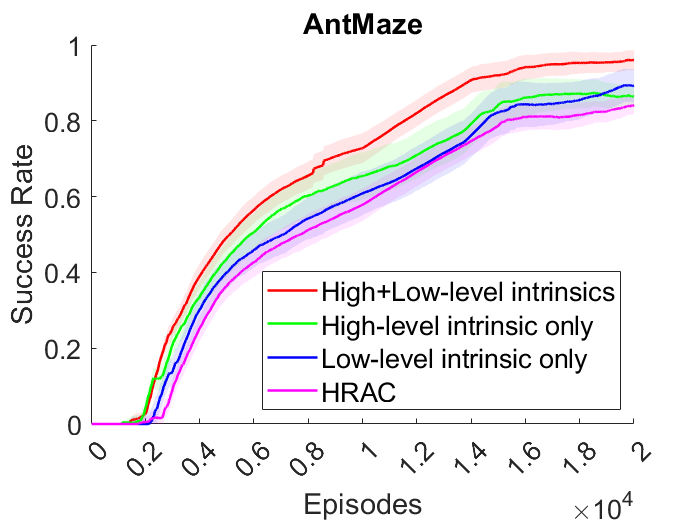}} 
    \subfigure[]{\includegraphics[width=0.32\textwidth]{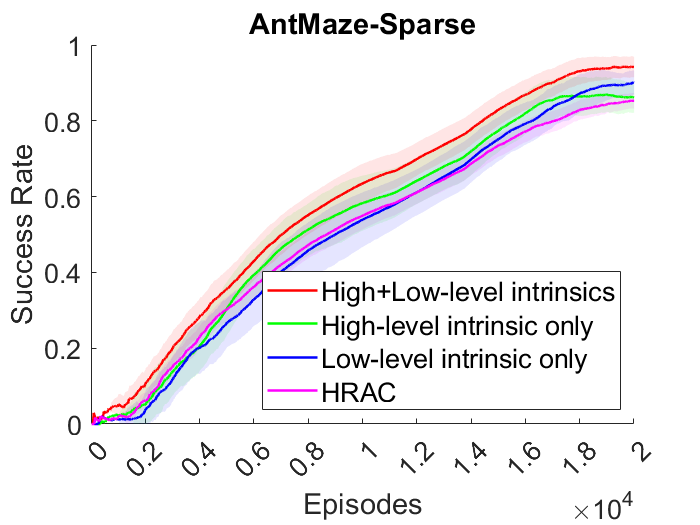}} 
    \subfigure[]{\includegraphics[width=0.32\textwidth]{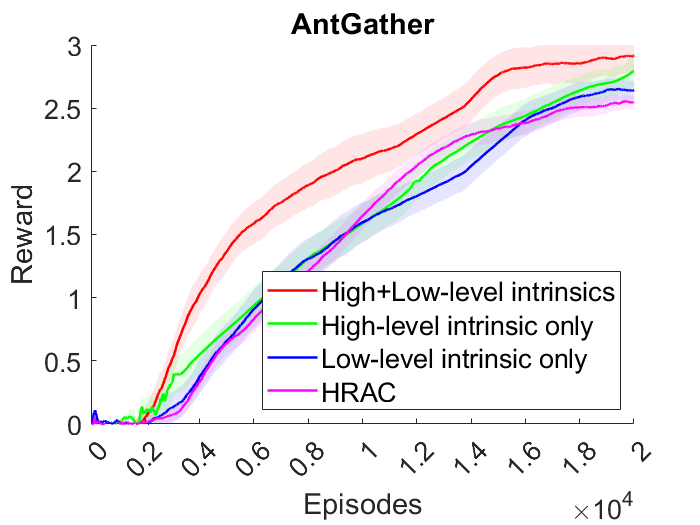}}
    \caption{Success Rate on (a) AntMaze (b) AntMaze-Sparse 
    and Reward on (c) AntGather using HRAC-G4RL, HRAC + High-level intrinsic, HRAC + Low-level intrinsic and HRAC. All curves have been smoothed equally for better visualization. The combination of high-level and low-level intrinsic rewards results in the highest success rates and fastest convergence.}
    \label{fig:HRAC_ab}
\end{figure}

\begin{figure}[h!]
    \centering
    \subfigure[]{\includegraphics[width=0.32\textwidth]{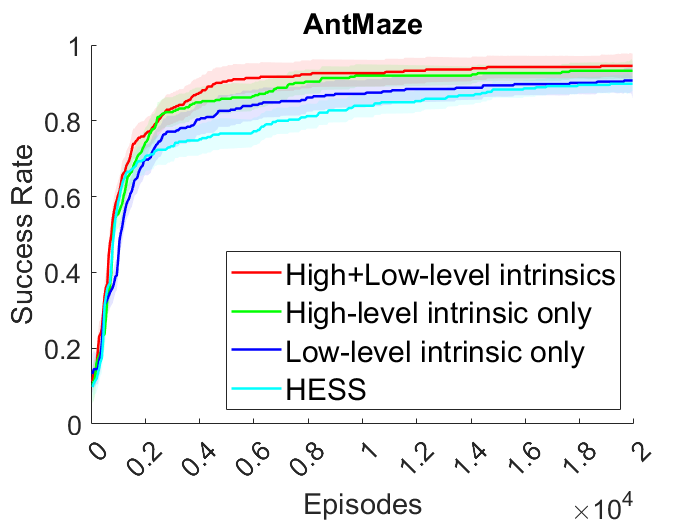}} 
    \subfigure[]{\includegraphics[width=0.32\textwidth]{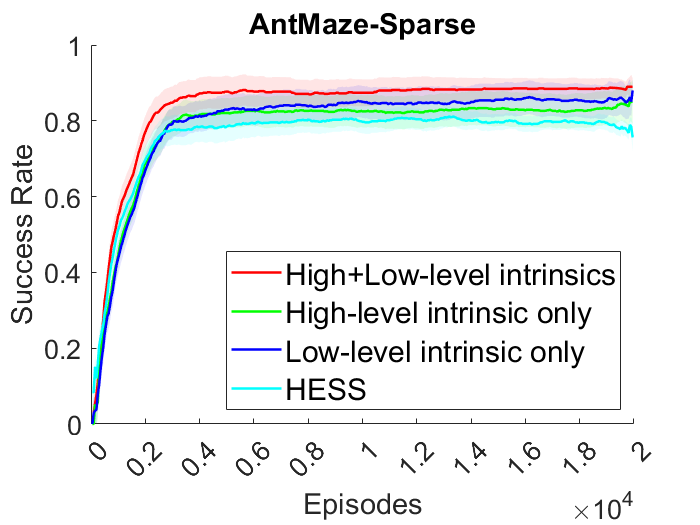}} 
    \subfigure[]{\includegraphics[width=0.32\textwidth]{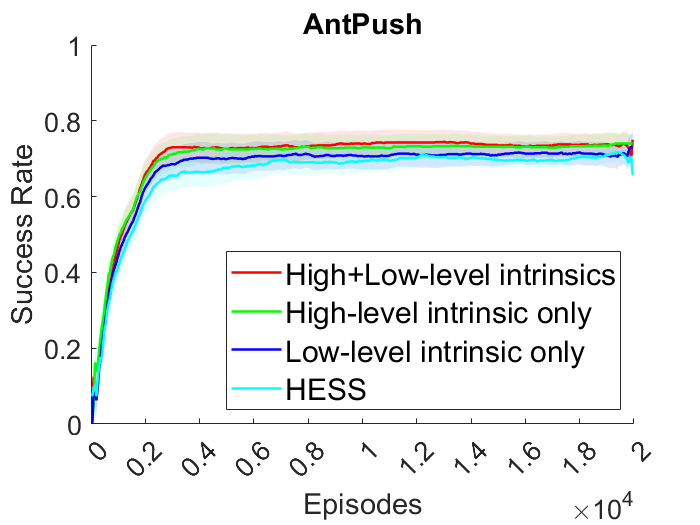}}
    \caption{Success Rate on (a) AntMaze (b) AntMaze-Sparse 
    and (c) AntPush using HESS-G4RL, HESS + High-level intrinsic, HESS + Low-level intrinsic and HESS. All curves have been equally smoothed for better visualization. The combination of high-level and low-level intrinsic rewards results in the highest success rates and fastest convergence.}
    \label{fig:HESS_ab}
\end{figure}

\begin{figure}[h!]
    \centering
    \subfigure[]{\includegraphics[width=0.32\textwidth]{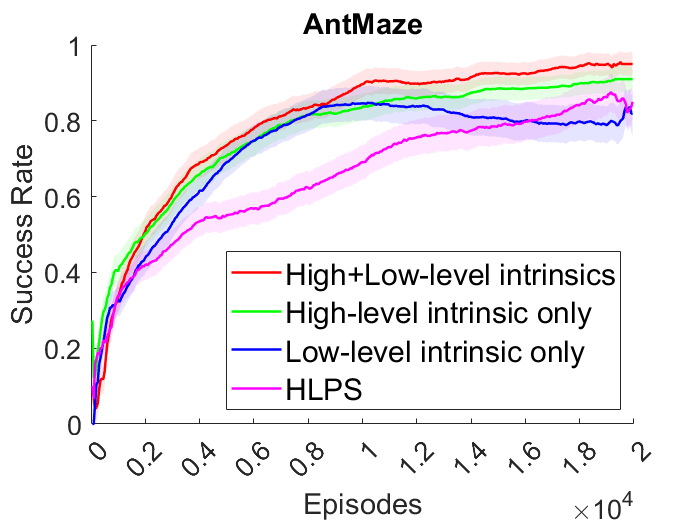}} 
    \subfigure[]{\includegraphics[width=0.32\textwidth]{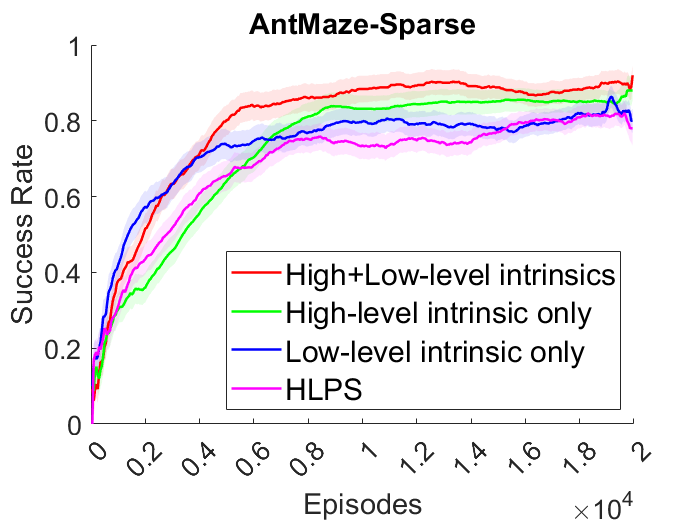}} 
    \subfigure[]{\includegraphics[width=0.32\textwidth]{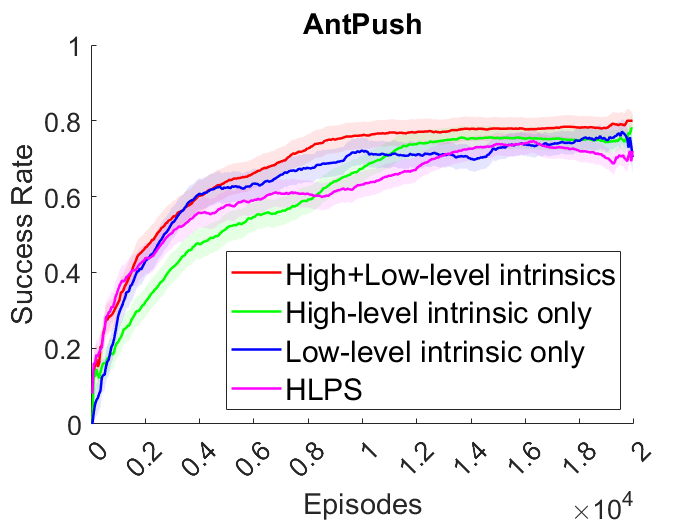}}
    \caption{Success Rate on (a) AntMaze (b) AntMaze-Sparse 
    and (c) AntPush using HLPS-G4RL, HLPS + High-level intrinsic, HLPS + Low-level intrinsic and HLPS. All curves have been equally smoothed for better visualization. The combination of high-level and low-level intrinsic rewards results in the highest success rates and fastest convergence.}
    \label{fig:HLPS_ab}
\end{figure}



Figures \ref{fig:HIRO_ab} to \ref{fig:HLPS_ab} show that, across all tested environments and algorithms, the combination of high-level and low-level intrinsic rewards results in the highest success rates and fastest convergence. The high-level intrinsic-only variant outperforms the low-level intrinsic-only variant, especially in sparse reward tasks, indicating that high-level intrinsic rewards play a crucial role in facilitating efficient exploration by encouraging the agent to select reachable and meaningful subgoals. In contrast, low-level intrinsic rewards have limited effect on exploration, primarily refining the execution of local behaviors. These results demonstrate that intrinsic rewards at different hierarchy levels serve complementary functions, and their combination yields superior performance.

\subsubsection{Balancing between time and performance}

\begin{figure}[h!]
    \centering
    \subfigure[]{\includegraphics[width=0.33\textwidth]{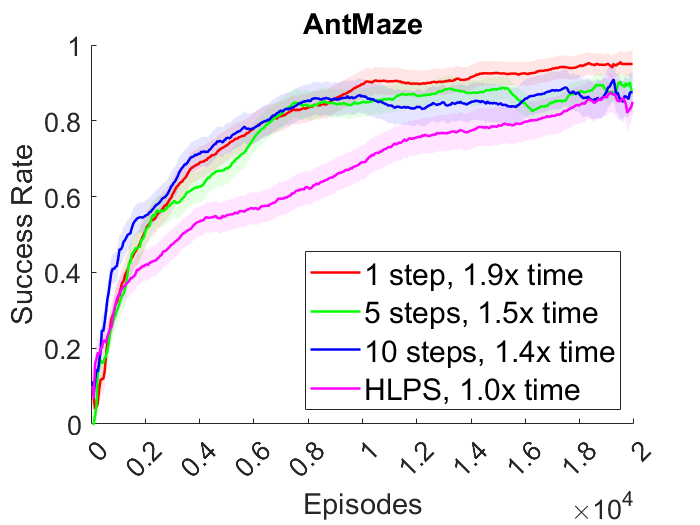}} 
    \subfigure[]{\includegraphics[width=0.33\textwidth]{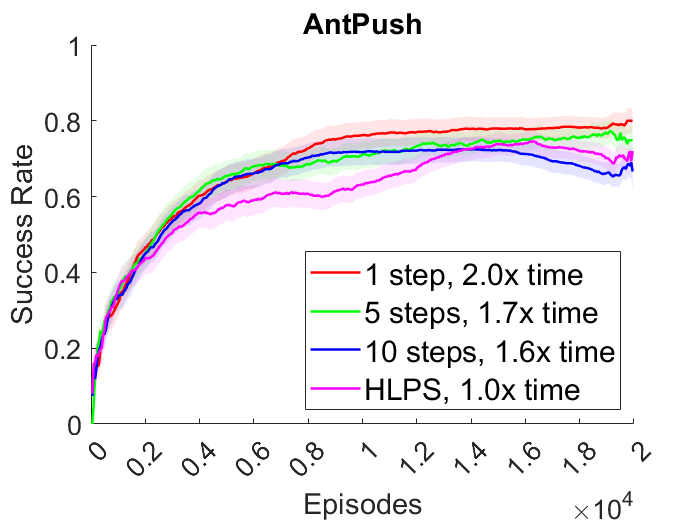}} 
    \caption{Success Rate on (a) AntMaze and (b) AntPush using HLPS+G4RL and HLPS. The number of steps in the legend indicates the selection of $t_c$ as described in Section \ref{ss:balance} and the timescale is calculated w.r.t. the vanila HLPS algorithm. Increasing the sampling interval substantially reduces computation time, with only minor degradation in success rates across both AntMaze and AntPush tasks.}
    \label{fig:HLPS_sample}
\end{figure}

\begin{figure}[h!]
    \centering
    \subfigure[]{\includegraphics[width=0.33\textwidth]{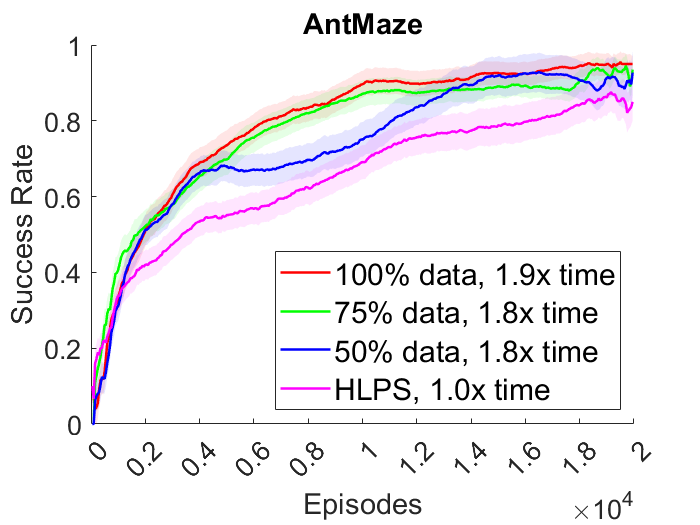}} 
    \subfigure[]{\includegraphics[width=0.33\textwidth]{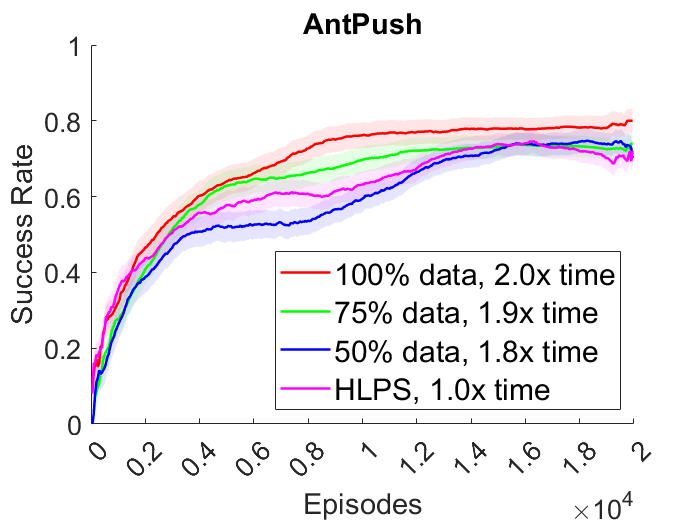}} 
    \caption{Success Rate on (a) AntMaze and (b) AntPush using HLPS+G4RL and HLPS. The percentage of data in the legend indicates the amount of data used in the training of the graph encoder-decoder as described in Section \ref{ss:balance} and the timescale is calculated w.r.t. the vanila HLPS algorithm. Reducing the amount of training data leads to only marginal improvements in computational efficiency and has negligible impact on final performance.}
    \label{fig:HLPS_data}
\end{figure}

To assess the trade-off between computational efficiency and performance, we evaluate two acceleration strategies mentioned in Section \ref{ss:balance}. First, we vary the sampling frequency of node features by testing intervals of 1, 5, and 10 steps in HLPS. As shown in Figure \ref{fig:HLPS_sample}, increasing the sampling interval substantially reduces computation time, as it decreases the number of interactions with the graph, with only minor degradation in success rates across both AntMaze and AntPush tasks. Second, we vary the proportion of training data used for the graph encoder-decoder, testing $50\%$, $75\%$, and $100\%$ subsets. Results in Figure \ref{fig:HLPS_data} indicate that reducing the amount of training data leads to only marginal improvements in computational efficiency and has negligible impact on final performance.

These findings suggest that the primary computational bottleneck of G4RL lies in the graph construction and node comparison processes described in Section \ref{sec:construction}, rather than in the encoder-decoder training itself. Adjusting the sampling frequency is therefore an effective approach for reducing time cost while largely preserving the benefits of G4RL integration.

\subsection{Subgoal space visualization}
\label{sec:vis}
This section shows how the subgoal space evolves in the AntMaze environment as the number of training episodes grows. We recorded state representations encountered in specific episodes
and then used the corresponding graph encoders from those episodes
to map these state representations 
to the subgoal representations. 
The subgoal representations are projected into 2D using t-SNE for visualization.

\begin{figure}[h!]
    \centering
    \subfigure[]{\includegraphics[width=0.3\textwidth]{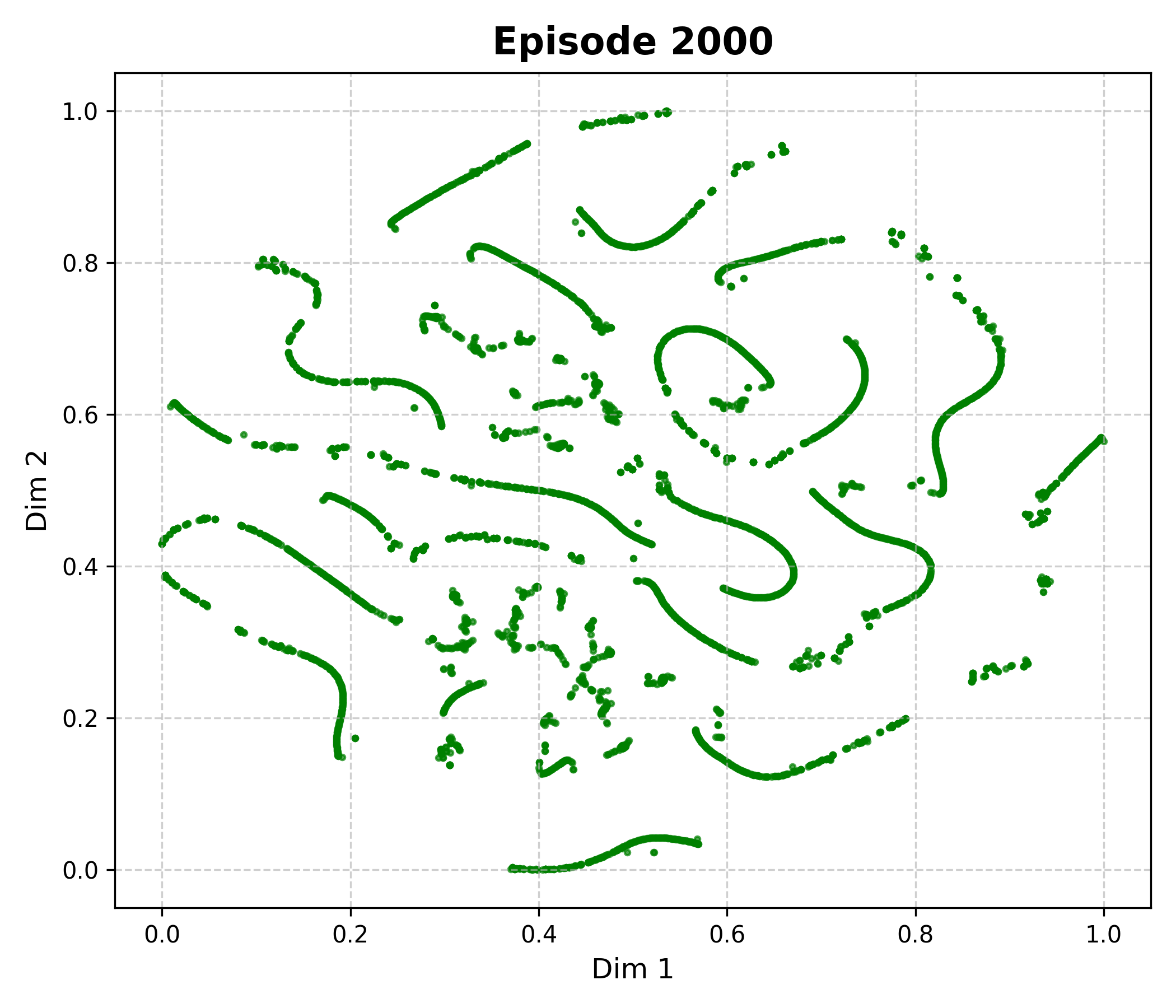}} 
    \subfigure[]{\includegraphics[width=0.3\textwidth]{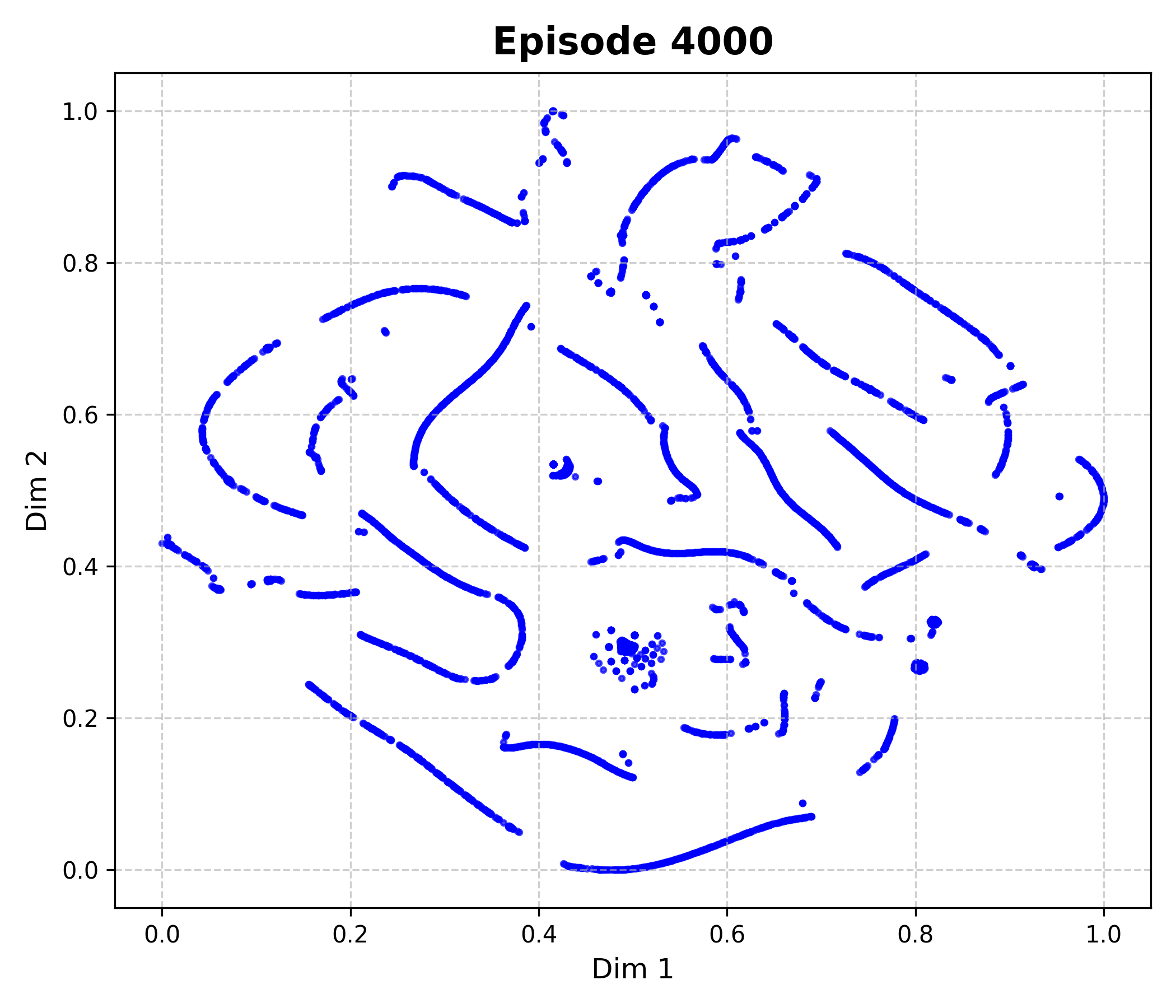}} 
    \subfigure[]{\includegraphics[width=0.3\textwidth]{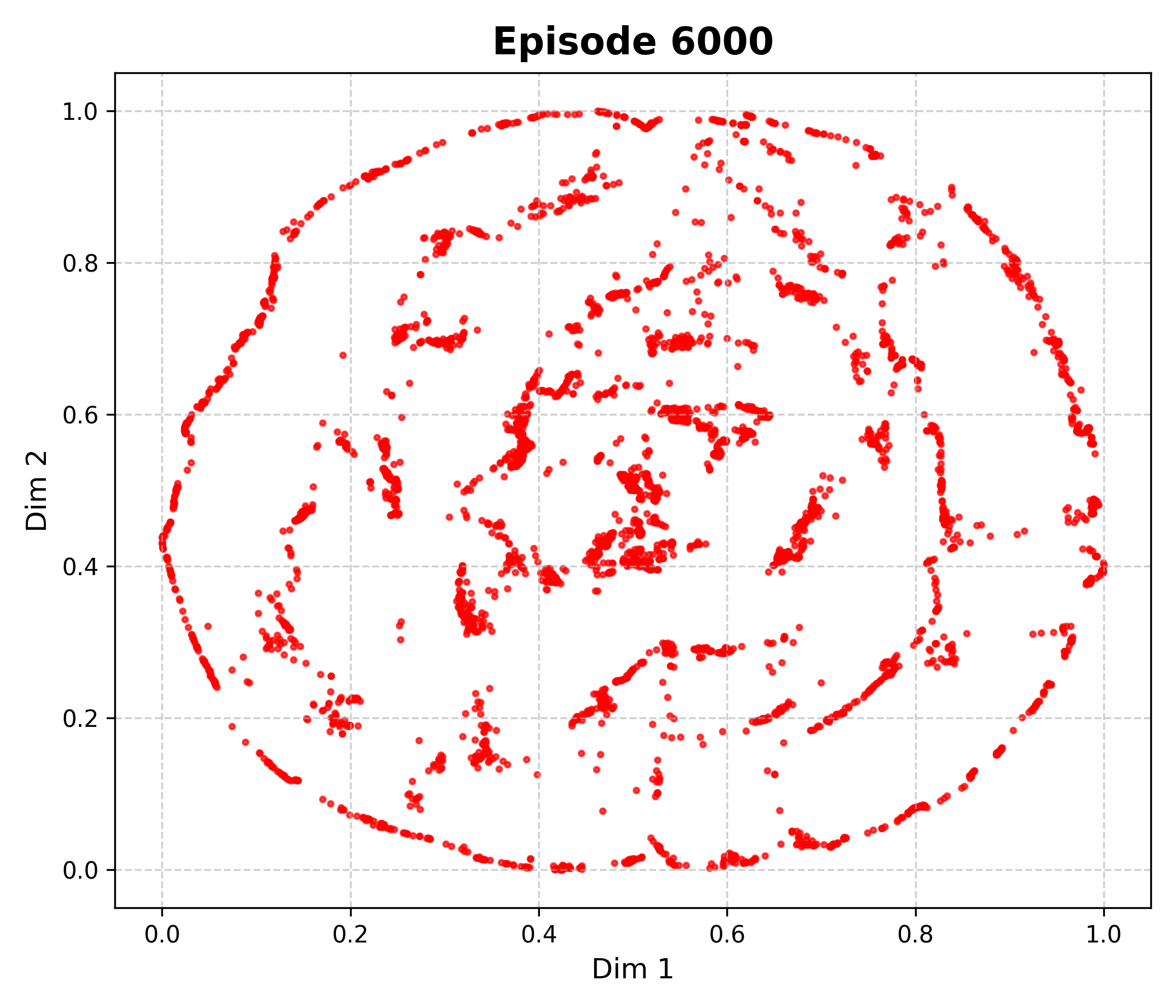}}
    \caption{The distribution of subgoal representations in the subgoal space projected by t-SNE at the (a) 2000th (b) 4000th (c) 6000th episode. The subgoal space evolves as the training comes into later episodes.}
    \label{fig:subgoal_dist}
\end{figure}

The distributions of subgoal representations in the subgoal space across different episodes are shown in Figure \ref{fig:subgoal_dist}.

From the figure we can conclude that the graph encoder-decoder gives better subgoal representations when the number of episodes grows. At the 2000th episode, the subgoal representations form a few chains in the 2D space, indicating that the current graph is partitioned into many isolated sub-graphs. The graph encoder has not yet learned to understand the environment well, and the agent cannot find relationships between different state representations yet. By the 4000th episode, the representations are less isolated, suggesting that the graph encoder has begun to learn the connections between different groups of state representation. Finally, at the 6000th episode, the subgoal representations are distributed more evenly across the entire space, indicating that the encoder can organize them into a highly connected graph. 
The subgoal space now better represents the underlying problem's structure.

The evolution of the graph encoder demonstrates that it can indeed learn to represent the task structure as the number of episodes grows.

\section{Conclusion}\label{s:con}
We have presented a novel approach using a graph encoder-decoder to address the challenges of poor subgoal representations and sample inefficiency in GCHRL.
The proposed architecture is designed to efficiently evaluate unseen states by operating in the graph representation space. It is easy to implement and can be seamlessly integrated into any existing GCHRL algorithms to enhance their performance in primarily symmetric environments. 
Our experiments on both sparse and dense control tasks have demonstrated the effectiveness and robustness of our method.

\bibliography{tmlr}

@inproceedings{ou2016asymmetric,
  title={Asymmetric transitivity preserving graph embedding},
  author={Ou, Mingdong and Cui, Peng and Pei, Jian and Zhang, Ziwei and Zhu, Wenwu},
  booktitle={Proceedings of the 22nd ACM SIGKDD international conference on Knowledge discovery and data mining},
  pages={1105--1114},
  year={2016}
}

@inproceedings{cao2015grarep,
  title={Grarep: Learning graph representations with global structural information},
  author={Cao, Shaosheng and Lu, Wei and Xu, Qiongkai},
  booktitle={Proceedings of the 24th ACM international on conference on information and knowledge management},
  pages={891--900},
  year={2015}
}

@inproceedings{ahmed2013distributed,
  title={Distributed large-scale natural graph factorization},
  author={Ahmed, Amr and Shervashidze, Nino and Narayanamurthy, Shravan and Josifovski, Vanja and Smola, Alexander J},
  booktitle={Proceedings of the 22nd international conference on World Wide Web},
  pages={37--48},
  year={2013}
}

@article{hong2022topological,
  title={Topological experience replay},
  author={Hong, Zhang-Wei and Chen, Tao and Lin, Yen-Chen and Pajarinen, Joni and Agrawal, Pulkit},
  journal={arXiv preprint arXiv:2203.15845},
  year={2022}
}

@article{zhu2022value,
  title={Value memory graph: A graph-structured world model for offline reinforcement learning},
  author={Zhu, Deyao and Li, Li Erran and Elhoseiny, Mohamed},
  journal={arXiv preprint arXiv:2206.04384},
  year={2022}
}

@inproceedings{hsu2024thought,
  title={Thought graph: Generating thought process for biological reasoning},
  author={Hsu, Chi-Yang and Cox, Kyle and Xu, Jiawei and Tan, Zhen and Zhai, Tianhua and Hu, Mengzhou and Pratt, Dexter and Chen, Tianlong and Hu, Ziniu and Ding, Ying},
  booktitle={Companion Proceedings of the ACM Web Conference 2024},
  pages={537--540},
  year={2024}
}

@inproceedings{zhao2022simulate,
  title={Simulate Human Thinking: Cognitive Knowledge Graph Reasoning for Complex Question Answering},
  author={Zhao, Hong and Fu, Yao and Jiang, Weihao and Pu, Shiliang and Cai, Xiaoyu},
  booktitle={Pacific-Asia Conference on Knowledge Discovery and Data Mining},
  pages={522--534},
  year={2022},
  organization={Springer}
}

@inproceedings{duan2016benchmarking,
  title={Benchmarking deep reinforcement learning for continuous control},
  author={Duan, Yan and Chen, Xi and Houthooft, Rein and Schulman, John and Abbeel, Pieter},
  booktitle={International conference on machine learning},
  pages={1329--1338},
  year={2016},
  organization={PMLR}
}

@article{luo2024goal,
  title={Goal-Conditioned Hierarchical Reinforcement Learning With High-Level Model Approximation},
  author={Luo, Yu and Ji, Tianying and Sun, Fuchun and Liu, Huaping and Zhang, Jianwei and Jing, Mingxuan and Huang, Wenbing},
  journal={IEEE Transactions on Neural Networks and Learning Systems},
  year={2024},
  publisher={IEEE}
}

@article{kim2021landmark,
  title={Landmark-guided subgoal generation in hierarchical reinforcement learning},
  author={Kim, Junsu and Seo, Younggyo and Shin, Jinwoo},
  journal={Advances in neural information processing systems},
  volume={34},
  pages={28336--28349},
  year={2021}
}

@article{eysenbach2019search,
  title={Search on the replay buffer: Bridging planning and reinforcement learning},
  author={Eysenbach, Ben and Salakhutdinov, Russ R and Levine, Sergey},
  journal={Advances in neural information processing systems},
  volume={32},
  year={2019}
}

@article{nehmzow2013novelty,
  title={Novelty detection as an intrinsic motivation for cumulative learning robots},
  author={Nehmzow, Ulrich and Gatsoulis, Yiannis and Kerr, Emmett and Condell, Joan and Siddique, Nazmul and McGuinnity, T Martin},
  journal={Intrinsically Motivated Learning in Natural and Artificial Systems},
  pages={185--207},
  year={2013},
  publisher={Springer}
}

@inproceedings{csimcsek2006intrinsic,
  title={An intrinsic reward mechanism for efficient exploration},
  author={Simsek, Ozgur and Barto, Andrew G},
  booktitle={Proceedings of the 23rd international conference on Machine learning},
  pages={833--840},
  year={2006}
}

@article{ha2018world,
  title={World models},
  author={Ha, David and Schmidhuber, J{\"u}rgen},
  journal={arXiv preprint arXiv:1803.10122},
  year={2018}
}

@inproceedings{zhang2021world,
  title={World model as a graph: Learning latent landmarks for planning},
  author={Zhang, Lunjun and Yang, Ge and Stadie, Bradly C},
  booktitle={International conference on machine learning},
  pages={12611--12620},
  year={2021},
  organization={PMLR}
}

@article{khoshraftar2024survey,
  title={A survey on graph representation learning methods},
  author={Khoshraftar, Shima and An, Aijun},
  journal={ACM Transactions on Intelligent Systems and Technology},
  volume={15},
  number={1},
  pages={1--55},
  year={2024},
  publisher={ACM New York, NY}
}

@article{chen2020graph,
  title={Graph representation learning: a survey},
  author={Chen, Fenxiao and Wang, Yun-Cheng and Wang, Bin and Kuo, C-C Jay},
  journal={APSIPA Transactions on Signal and Information Processing},
  volume={9},
  pages={e15},
  year={2020},
  publisher={Cambridge University Press}
}

@article{gieselmann2021planning,
  title={Planning-augmented hierarchical reinforcement learning},
  author={Gieselmann, Robert and Pokorny, Florian T},
  journal={IEEE Robotics and Automation Letters},
  volume={6},
  number={3},
  pages={5097--5104},
  year={2021},
  publisher={IEEE}
}

@article{klissarov2023deep,
  title={Deep laplacian-based options for temporally-extended exploration},
  author={Klissarov, Martin and Machado, Marlos C},
  journal={arXiv preprint arXiv:2301.11181},
  year={2023}
}

@inproceedings{wan2021reasoning,
  title={Reasoning like human: Hierarchical reinforcement learning for knowledge graph reasoning},
  author={Wan, Guojia and Pan, Shirui and Gong, Chen and Zhou, Chuan and Haffari, Gholamreza},
  booktitle={International Joint Conference on Artificial Intelligence},
  year={2021},
  organization={International Joint Conference on Artificial Intelligence}
}

@article{shang2019learning,
  title={Learning world graphs to accelerate hierarchical reinforcement learning},
  author={Shang, Wenling and Trott, Alex and Zheng, Stephan and Xiong, Caiming and Socher, Richard},
  journal={arXiv preprint arXiv:1907.00664},
  year={2019}
}

@article{lee2022dhrl,
  title={DHRL: a graph-based approach for long-horizon and sparse hierarchical reinforcement learning},
  author={Lee, Seungjae and Kim, Jigang and Jang, Inkyu and Kim, H Jin},
  journal={Advances in Neural Information Processing Systems},
  volume={35},
  pages={13668--13678},
  year={2022}
}

@article{kulkarni2016hierarchical,
  title={Hierarchical deep reinforcement learning: Integrating temporal abstraction and intrinsic motivation},
  author={Kulkarni, Tejas D and Narasimhan, Karthik and Saeedi, Ardavan and Tenenbaum, Josh},
  journal={Advances in neural information processing systems},
  volume={29},
  year={2016}
}

@inproceedings{vezhnevets2017feudal,
  title={Feudal networks for hierarchical reinforcement learning},
  author={Vezhnevets, Alexander Sasha and Osindero, Simon and Schaul, Tom and Heess, Nicolas and Jaderberg, Max and Silver, David and Kavukcuoglu, Koray},
  booktitle={International conference on machine learning},
  pages={3540--3549},
  year={2017},
  organization={PMLR}
}

@article{levy2017learning,
  title={Learning multi-level hierarchies with hindsight},
  author={Levy, Andrew and Konidaris, George and Platt, Robert and Saenko, Kate},
  journal={arXiv preprint arXiv:1712.00948},
  year={2017}
}

@article{nachum2018data,
  title={Data-efficient hierarchical reinforcement learning},
  author={Nachum, Ofir and Gu, Shixiang Shane and Lee, Honglak and Levine, Sergey},
  journal={Advances in neural information processing systems},
  volume={31},
  year={2018}
}

@article{nachum2018near,
  title={Near-optimal representation learning for hierarchical reinforcement learning},
  author={Nachum, Ofir and Gu, Shixiang and Lee, Honglak and Levine, Sergey},
  journal={arXiv preprint arXiv:1810.01257},
  year={2018}
}

@article{huang2019mapping,
  title={Mapping state space using landmarks for universal goal reaching},
  author={Huang, Zhiao and Liu, Fangchen and Su, Hao},
  journal={Advances in Neural Information Processing Systems},
  volume={32},
  year={2019}
}

@article{wang2024probabilistic,
  title={Probabilistic Subgoal Representations for Hierarchical Reinforcement learning},
  author={Wang, Vivienne Huiling and Wang, Tinghuai and Yang, Wenyan and K{\"a}m{\"a}r{\"a}inen, Joni-Kristian and Pajarinen, Joni},
  journal={arXiv preprint arXiv:2406.16707},
  year={2024}
}

@article{guo2021state,
  title={State-temporal compression in reinforcement learning with the reward-restricted geodesic metric},
  author={Guo, Shangqi and Yan, Qi and Su, Xin and Hu, Xiaolin and Chen, Feng},
  journal={IEEE Transactions on Pattern Analysis and Machine Intelligence},
  volume={44},
  number={9},
  pages={5572--5589},
  year={2021},
  publisher={IEEE}
}

@article{zhang2022adjacency,
  title={Adjacency constraint for efficient hierarchical reinforcement learning},
  author={Zhang, Tianren and Guo, Shangqi and Tan, Tian and Hu, Xiaolin and Chen, Feng},
  journal={IEEE Transactions on Pattern Analysis and Machine Intelligence},
  volume={45},
  number={4},
  pages={4152--4166},
  year={2022},
  publisher={IEEE}
}

@inproceedings{todorov2012mujoco,
  title={Mujoco: A physics engine for model-based control},
  author={Todorov, Emanuel and Erez, Tom and Tassa, Yuval},
  booktitle={2012 IEEE/RSJ international conference on intelligent robots and systems},
  pages={5026--5033},
  year={2012},
  organization={IEEE}
}

@book{puterman2014markov,
  title={Markov decision processes: discrete stochastic dynamic programming},
  author={Puterman, Martin L},
  year={2014},
  publisher={John Wiley \& Sons}
}

@article{sutton1999between,
  title={Between MDPs and semi-MDPs: A framework for temporal abstraction in reinforcement learning},
  author={Sutton, Richard S and Precup, Doina and Singh, Satinder},
  journal={Artificial intelligence},
  volume={112},
  number={1-2},
  pages={181--211},
  year={1999},
  publisher={Elsevier}
}

@inproceedings{fujimoto2018addressing,
  title={Addressing function approximation error in actor-critic methods},
  author={Fujimoto, Scott and Hoof, Herke and Meger, David},
  booktitle={International conference on machine learning},
  pages={1587--1596},
  year={2018},
  organization={PMLR}
}

@article{haarnoja2018soft,
  title={Soft actor-critic algorithms and applications},
  author={Haarnoja, Tuomas and Zhou, Aurick and Hartikainen, Kristian and Tucker, George and Ha, Sehoon and Tan, Jie and Kumar, Vikash and Zhu, Henry and Gupta, Abhishek and Abbeel, Pieter and others},
  journal={arXiv preprint arXiv:1812.05905},
  year={2018}
}

@article{schulman2017proximal,
  title={Proximal policy optimization algorithms},
  author={Schulman, John and Wolski, Filip and Dhariwal, Prafulla and Radford, Alec and Klimov, Oleg},
  journal={arXiv preprint arXiv:1707.06347},
  year={2017}
}

@article{klissarov2020reward,
  title={Reward propagation using graph convolutional networks},
  author={Klissarov, Martin and Precup, Doina},
  journal={Advances in Neural Information Processing Systems},
  volume={33},
  pages={12895--12908},
  year={2020}

}

@article{hamilton2020graph,
  title={Graph representation learning},
  author={Hamilton, William L},
  journal={Synthesis Lectures on Artifical Intelligence and Machine Learning},
  volume={14},
  number={3},
  pages={1--159},
  year={2020},
  publisher={Morgan \& Claypool Publishers}
}

@article{kingma2014adam,
  title={Adam: A method for stochastic optimization},
  author={Kingma, Diederik P and Ba, Jimmy},
  journal={arXiv preprint arXiv:1412.6980},
  year={2014}
}

@article{li2022active,
  title={Active hierarchical exploration with stable subgoal representation learning},
  author={Li, Siyuan and Zhang, Jin and Wang, Jianhao and Yu, Yang and Zhang, Chongjie},
  journal={arXiv preprint arXiv:2105.14750},
  year={2022}
}
\bibliographystyle{tmlr}

\appendix
\newpage
\section{Limitations and Future work}
Despite the advantages demonstrated by G4RL in HRL tasks, its effectiveness depends significantly on several hyperparameters (e.g., $\epsilon_d$, $\alpha_l$, and $\alpha_h$), which require careful tuning to achieve optimal performance across different environments.

In future work, we aim to develop methods for automatically selecting these hyperparameters based on environmental dynamics, thereby reducing the need for manual tuning. Also, we plan to extend our work by exploring how to generate subgoals with more interpretable representations to facilitate knowledge transfer, potentially leveraging alternative graph representations (e.g. Graph Laplacian). Another promising direction is to transfer the knowledge embedded in the graph structure to new tasks by analyzing graph topology and establishing mappings between nodes of different state graphs. Additionally, more evidence on asymmetric environments is needed to demonstrate G4RL's robustness when asymmetric/irreversible transitions are present.

\newpage
\section{Implementation details}
\subsection{Environment details}

\textbf{AntMaze} This environment is a part of the Gymnasium-Robotics libraries. The size of the environment is $24 \times 24$. Both of the state space and the action space are continuous, with a state dimension of 31 and an action dimension of 8. The reward of each step is defined by its negative Euclidean distance from the current location to the target position. At evaluation time, the goal is set to $(0, 16)$ and an episode is recognized as successful if the agent is within an Euclidean distance of 5 from the goal.

\textbf{AntMaze Sparse} This environment is a variant of Antmaze. The size of the environment is $20 \times 20$. The state and action spaces are the same as those in AntMaze. The reward of each step is 1 if and only if the agent reaches within an Euclidean distance of 1 from the goal, which is set to $(2,9)$.

\textbf{AntGather} This environment is described in \cite{duan2016benchmarking}. The size of the environment is $20 \times 20$. The state and action spaces are continuous. The task involves gathering apples to the designated place. The agent will be awarded $+1$ for each apple gathered and $-1$ for each bomb gathered. Apples and bombs are randomly placed in the $20 \times 20$ world.

\textbf{AntPush} The size of the environment is $20 \times 20$. The state and action spaces are continuous. A challenging task that requires both task and motion planning. The agent needs to move to the left then move up and push the block to the right in order to reach the goal.
\subsection{Network architecture details}
Our network architecture for the HRL agents is the same as described in \cite{nachum2018data}, \cite{zhang2022adjacency}, \cite{li2022active} and \cite{wang2024probabilistic}. For HIRO, HRAC, HESS and HLPS both the high-level and low-level agents use the TD3 algorithm. The size of each hidden layer in actor and critic networks in TD3 is $300$.

For the graph encoder-decoder, we use a four-layer fully connected network with the hidden size of $128$. The activation function used is ReLU. The decoder is a dot product of the two input subgoal representations.

We use Adam as the optimizer for the actor network, critic network and the graph encoder is Adam optimizer \citep{kingma2014adam}.
\newpage
\section{Hyperparameters}
In this section we list all hyperparameters used in our experiments.
\begin{table}[h!]
    \centering
    \begin{tabular}{cc}
        \textbf{Hyperparameters} & \textbf{Values} \\
        \hline\hline
         High-level agent& \\
         \hline
         Actor learning rate & 0.0001\\
         Critic learning rate & 0.001\\
         Batch size & 128\\
         Discount factor $\gamma$ & 0.99\\
         Policy update frequency & 1\\
         High-level action frequency & 10\\
         Replay buffer size & 20000 \\
         Exploration strategy & Gaussian($\sigma = 1$)\\
         \hline\hline
         Low-level agent& \\
         \hline
         Actor learning rate & 0.0001\\
         Critic learning rate & 0.001\\
         Batch size & 128\\
         Discount factor $\gamma$ & 0.99\\
         Policy update frequency & 1\\
         Replay buffer size & 20000 \\
         Exploration strategy & Gaussian($\sigma = 1$)\\
         \hline
    \end{tabular}
    \caption{Hyperparameters used in high- and low-level TD3 agents.}
    \label{tab:hyp1}
\end{table}
\begin{table}[h!]
    \centering
    \begin{tabular}{cc}
        \textbf{Hyperparameters} & \textbf{Values} \\
        \hline\hline
         Number of nodes $N$ & 200\\
         Batch size & 128\\
         Optimizer learning rate & 0.0001\\
         $\epsilon_d$ & 0.1 for AntMaze/0.2 for others\\
         $\alpha_h$ & 0.1\\
         $\alpha_l$ & 0.1 \\
         $\beta$ & 0.2\\
        \hline
    \end{tabular}
    \caption{Hyperparameters used in the graph encoder-decoder.}
    \label{tab:hyp2}
\end{table}
\newpage
\section{Algorithm}
\label{sec:algo}
Now we describe our proposed method in Algorithm \ref{alg:G4RL}.
\begin{algorithm} 
    \caption{GCHRL+G4RL} \label{alg:G4RL}
    \begin{algorithmic}[1]
    \Require 
    
    \text{High-level policy} $\pi_h(g|\phi(s))$, \text{low-level policy} $\pi_l(a|\phi(s), g)$, \text{replay buffer} $\mathcal{B}$,
    
    \text{graph encoder} $\mathbf{E}$, \text{graph decoder} $\mathbf{D}$, \text{high-level action frequency} $K$, 
    
    \text{significance hyperparameter} $\alpha_h$ and $\alpha_l$, \text{tolerance hyperparameter} $\beta$, 
    
    \text{number of episodes} $N$, \text{nunber of steps in one episode} $T$.

    \State $n = 0$
    \While{$n \leq N$}
    \State $t = 0$
    \State $c = 0$
    \While{$t \leq T$}

    \If {$t \mod K = 0$}
    \State Execute the high-level policy $\pi_h(g_t|\phi(s_t))$ to sample the subgoal $g_t$.
    \Else 
    \State Keep the subgoal $g_t$ unchanged.
    \EndIf
    \State Execute the low-level policy $\pi_l(a_t|\phi(s_t), g_t)$ to sample the atomic action $a_t$.

    \State Sample reward $r_t$ and next state $s_{t+1}$.

    \State Calculate $r_h(s_t, g_t, s_{t+1})$ and $r_l(s_t, g_t, a_t, s_{t+1})$ using 
    (\ref{eq:hir}) and (\ref{eq:lir}).
    
    \State Collect experience $(s_t, g_t, a_t, r_h, r_l, s_{t+1})$ and update the replay buffer $\mathcal{B}$.

    \State Update node representations and edge weights using collected experience.

    \State Update $c$ using (\ref{eq:updateschedule}).

    \If {$c \geq \beta$}
    
    \State Update graph encoder $\mathbf{E}$ with node representation and edge information in the graph.

    \State $c = 0$.
    \EndIf

    \State $t = t+1$

    \EndWhile

    \State Update low-level policy $\pi_l(a|\phi(s), g)$ using the chosen HRL algorithm.

    \State Update high-level policy $\pi_h(g|\phi(s))$ using the chosen HRL algorithm.

    \State $n = n+1$

    \EndWhile
    \end{algorithmic}
\end{algorithm}

\newpage
\section{Un-smoothed experiment results}
\label{sec:unsmooth}
In this section, we present the un-smoothed versions of figures 4 and 5 in the main text to show the full details of the experiments.
\begin{figure}[h!]
    \centering
    \subfigure[]{\includegraphics[width=0.32\textwidth]{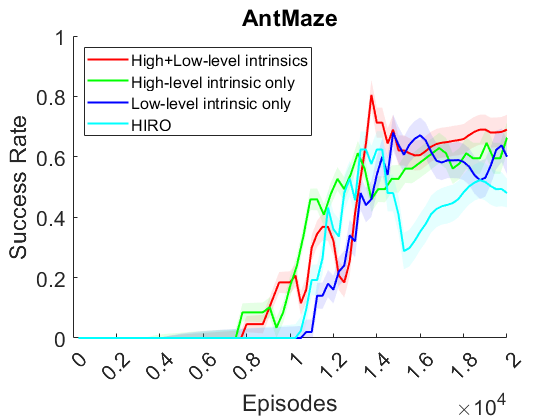}} 
    \subfigure[]{\includegraphics[width=0.32\textwidth]{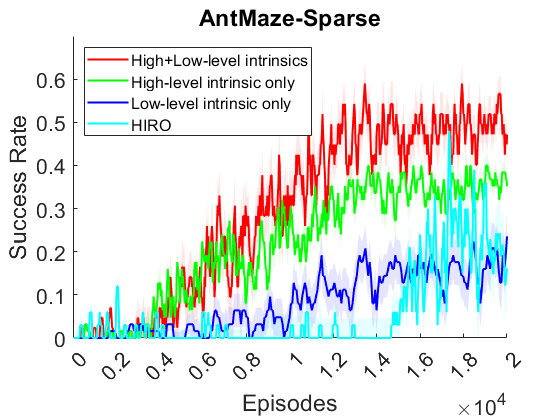}} 
    \subfigure[]{\includegraphics[width=0.32\textwidth]{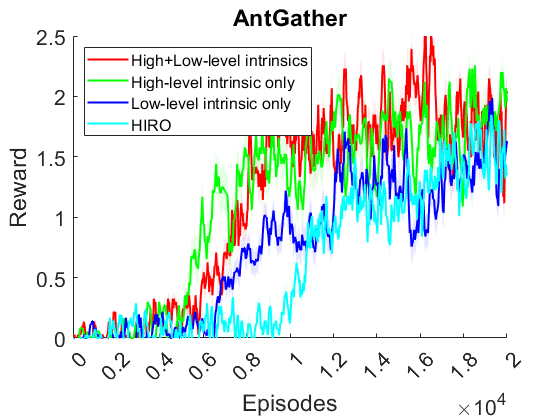}}
    \caption{Success Rate on (a) AntMaze (b) AntMaze-Sparse 
    and Reward on (c) AntGather using HIRO-G4RL, HIRO + High-level intrinsic, HIRO + Low-level intrinsic and HIRO. Smoothing is removed in contrast to figure \ref{fig:HIRO_ab}}
    \label{fig:HIRO_ab_raw}
\end{figure}

\begin{figure}[h!]
    \centering
    \subfigure[]{\includegraphics[width=0.32\textwidth]{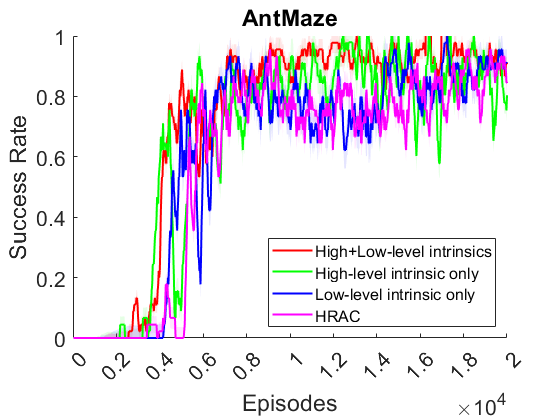}} 
    \subfigure[]{\includegraphics[width=0.32\textwidth]{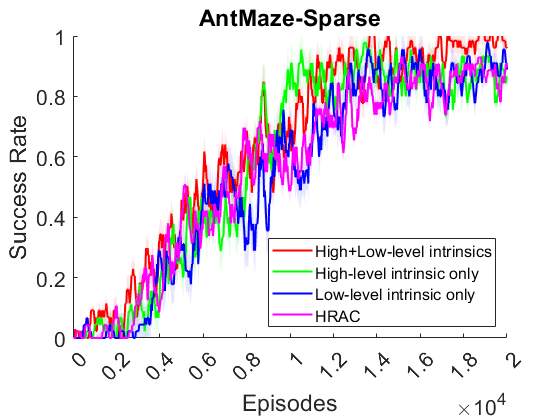}} 
    \subfigure[]{\includegraphics[width=0.32\textwidth]{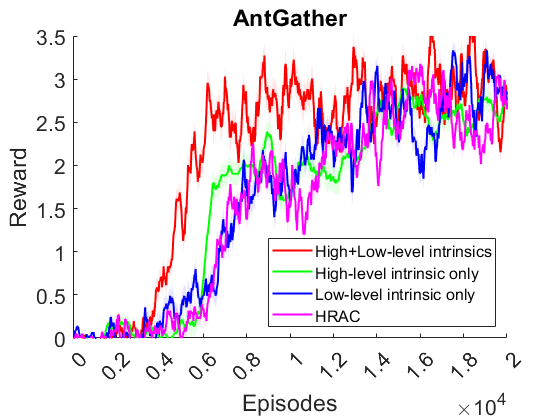}}
    \caption{Success Rate on (a) AntMaze (b) AntMaze-Sparse 
    and Reward on (c) AntGather using HRAC-G4RL, HRAC + High-level intrinsic, HRAC + Low-level intrinsic and HRAC. Smoothing is removed in contrast to figure \ref{fig:HRAC_ab}}
    \label{fig:HRAC_ab_raw}
\end{figure}

\newpage
\section{Exploring the subgoal space through diverse visualization techniques}
\begin{figure}[h!]
    \centering
    \subfigure[]{\includegraphics[width=0.3\textwidth]{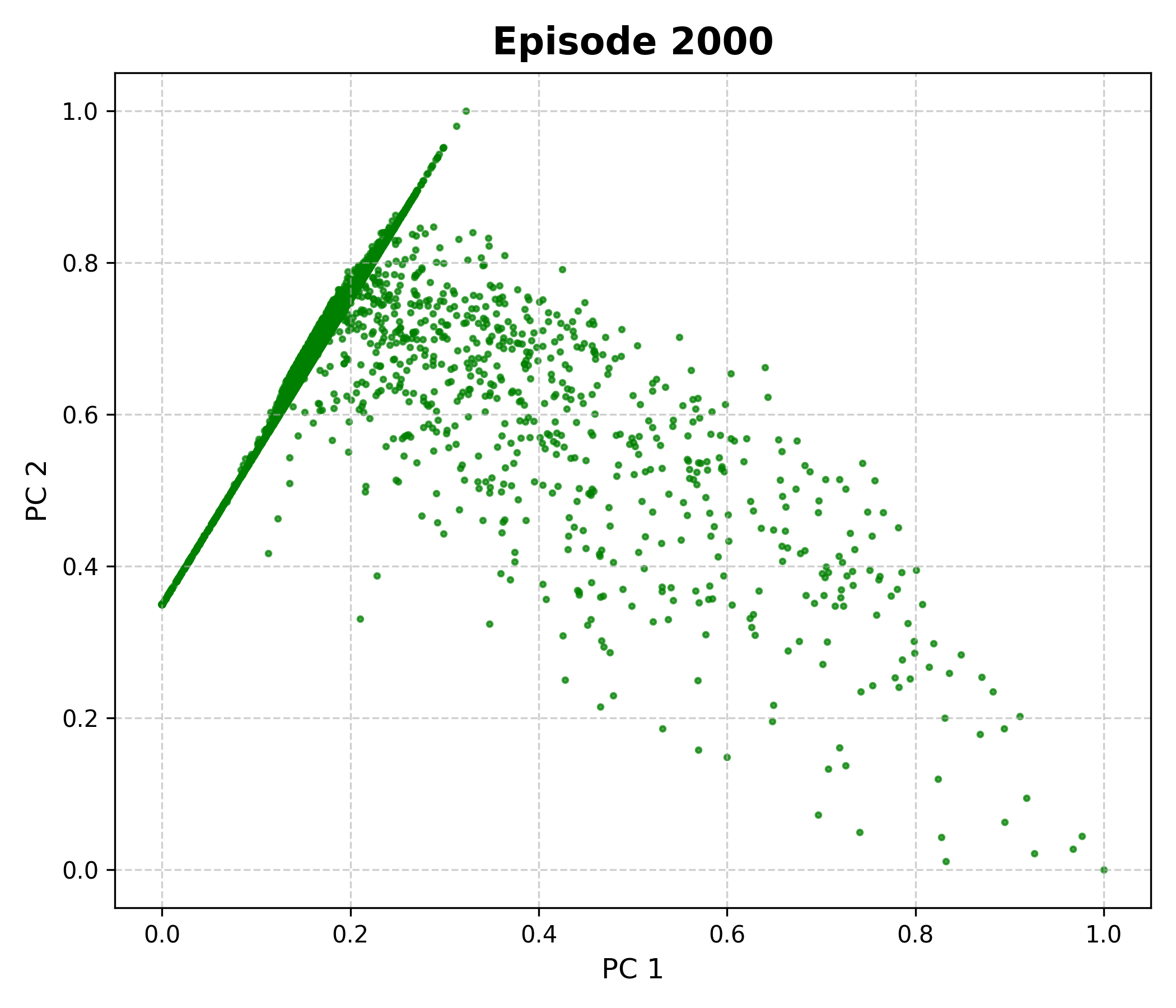}} 
    \subfigure[]{\includegraphics[width=0.3\textwidth]{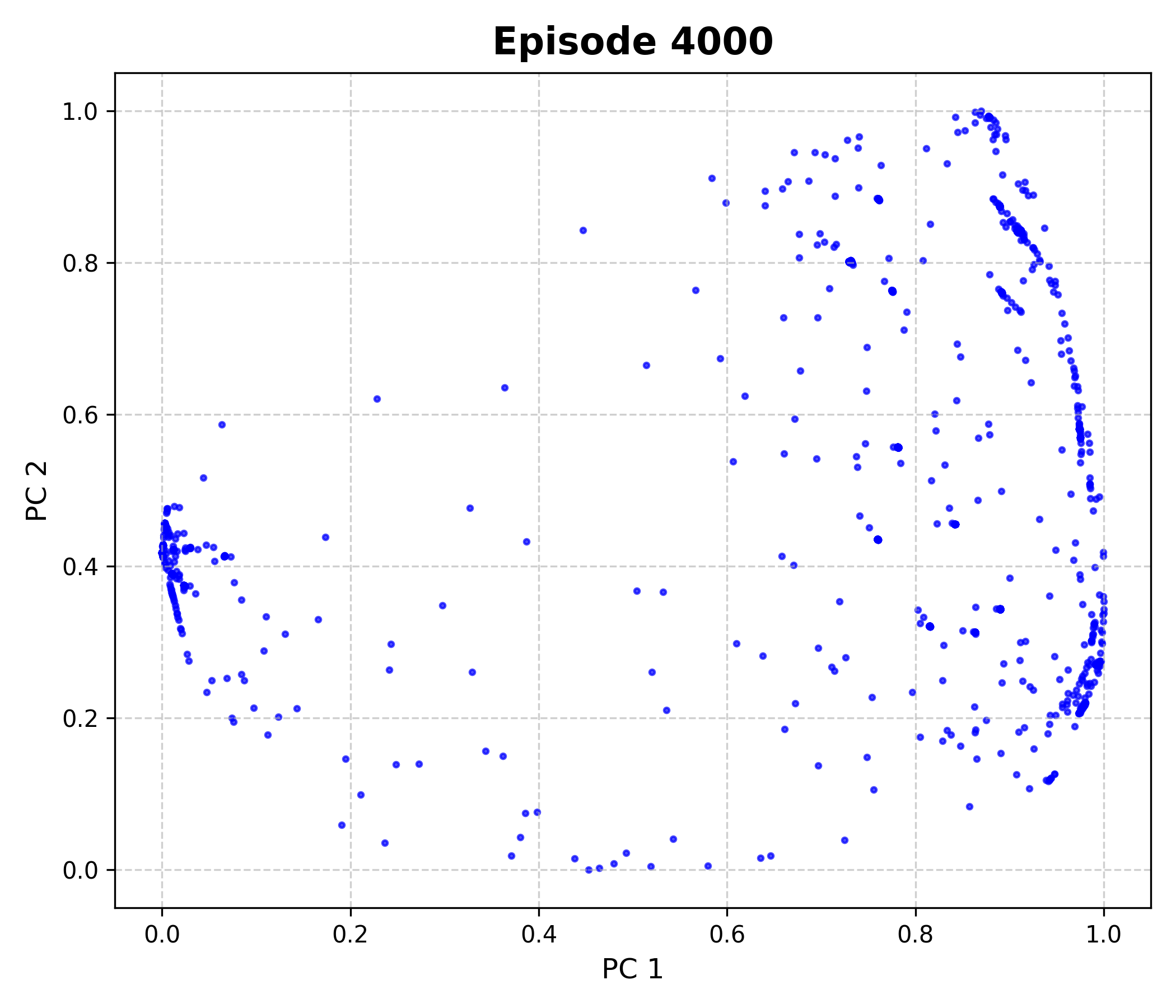}} 
    \subfigure[]{\includegraphics[width=0.3\textwidth]{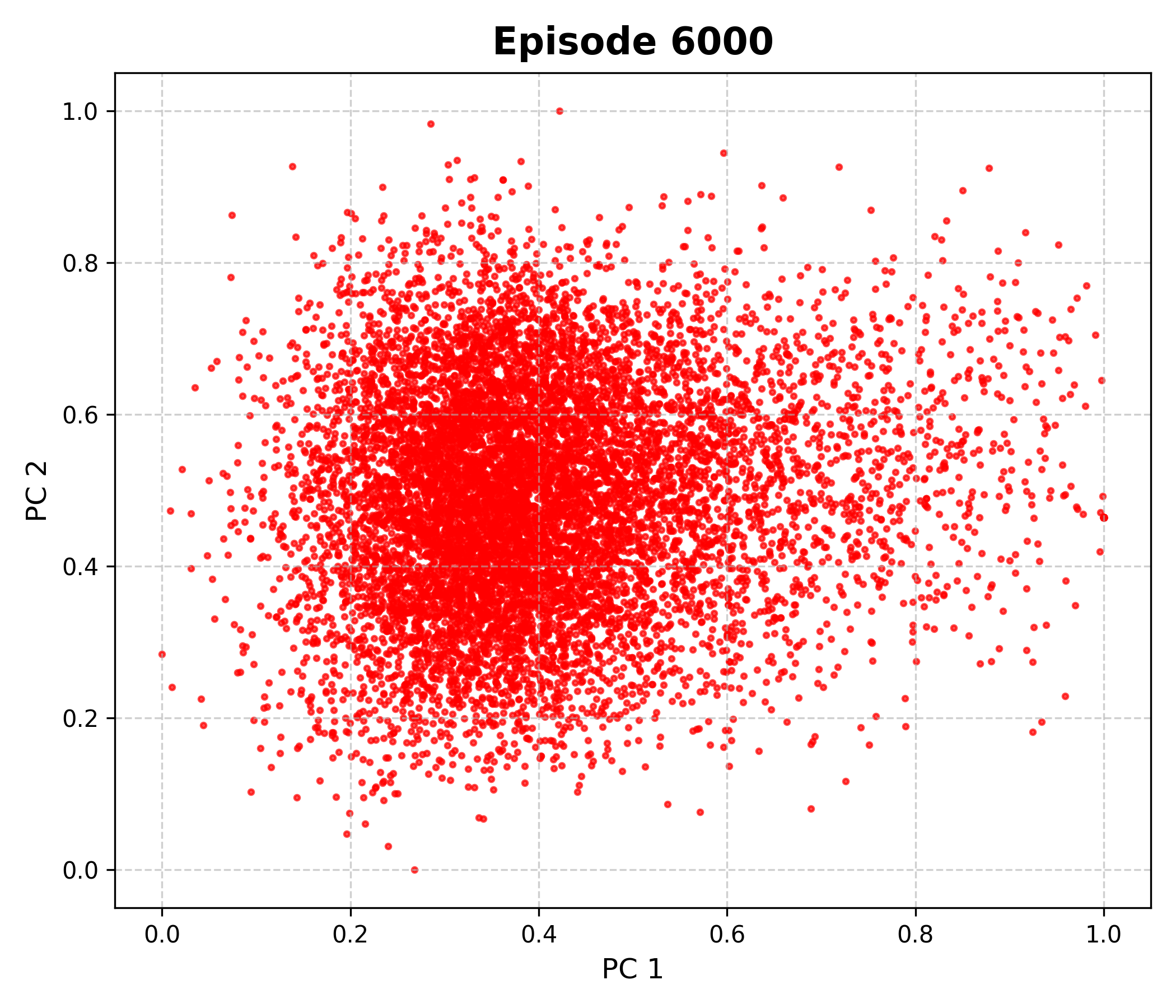}}
    \caption{The distribution of subgoal representations in the subgoal space projected by PCA at the (a) 2000th (b) 4000th (c) 6000th episode.}
    \label{fig:subgoal_dist_pca}
\end{figure}

\begin{figure}[h!]
    \centering
    \subfigure[]{\includegraphics[width=0.3\textwidth]{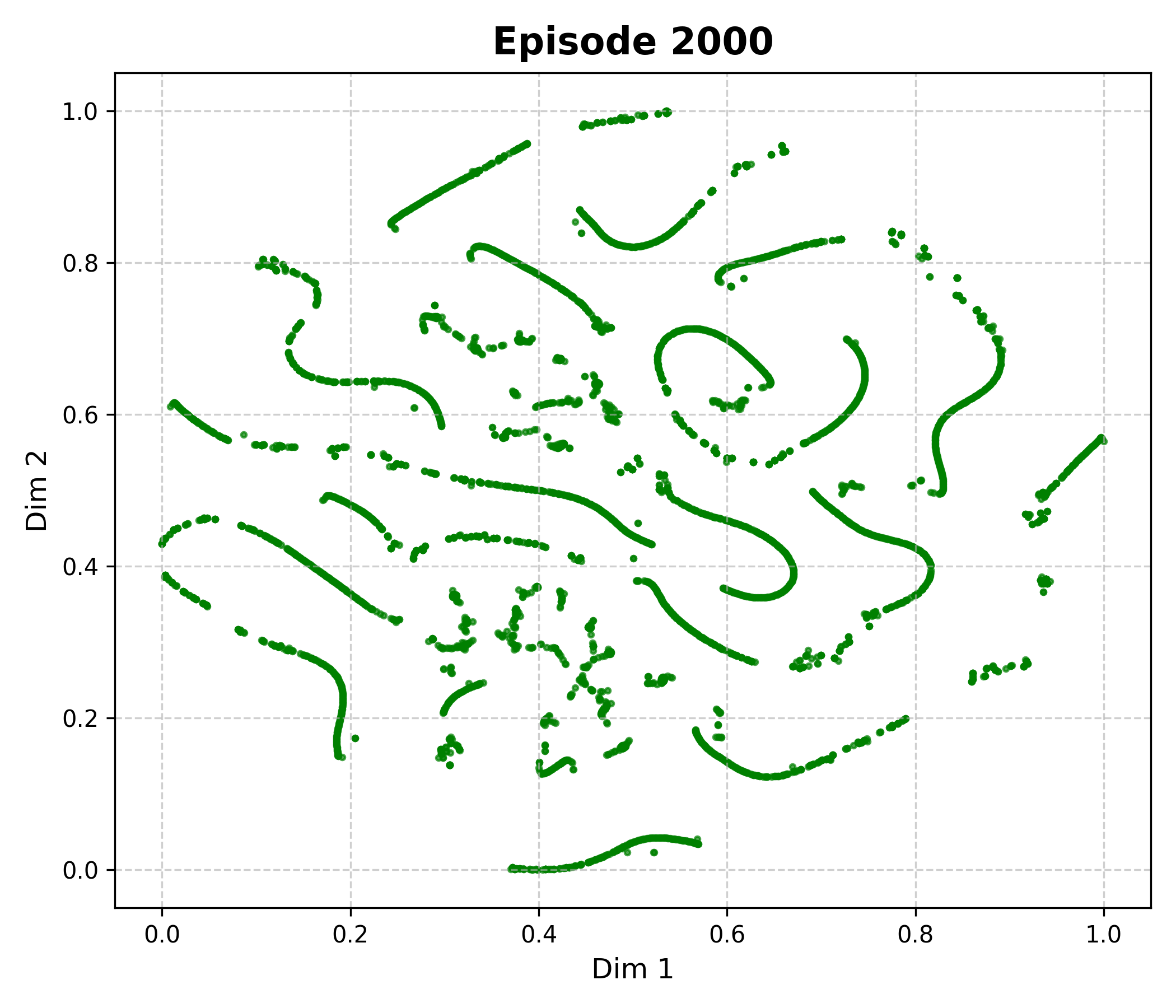}} 
    \subfigure[]{\includegraphics[width=0.3\textwidth]{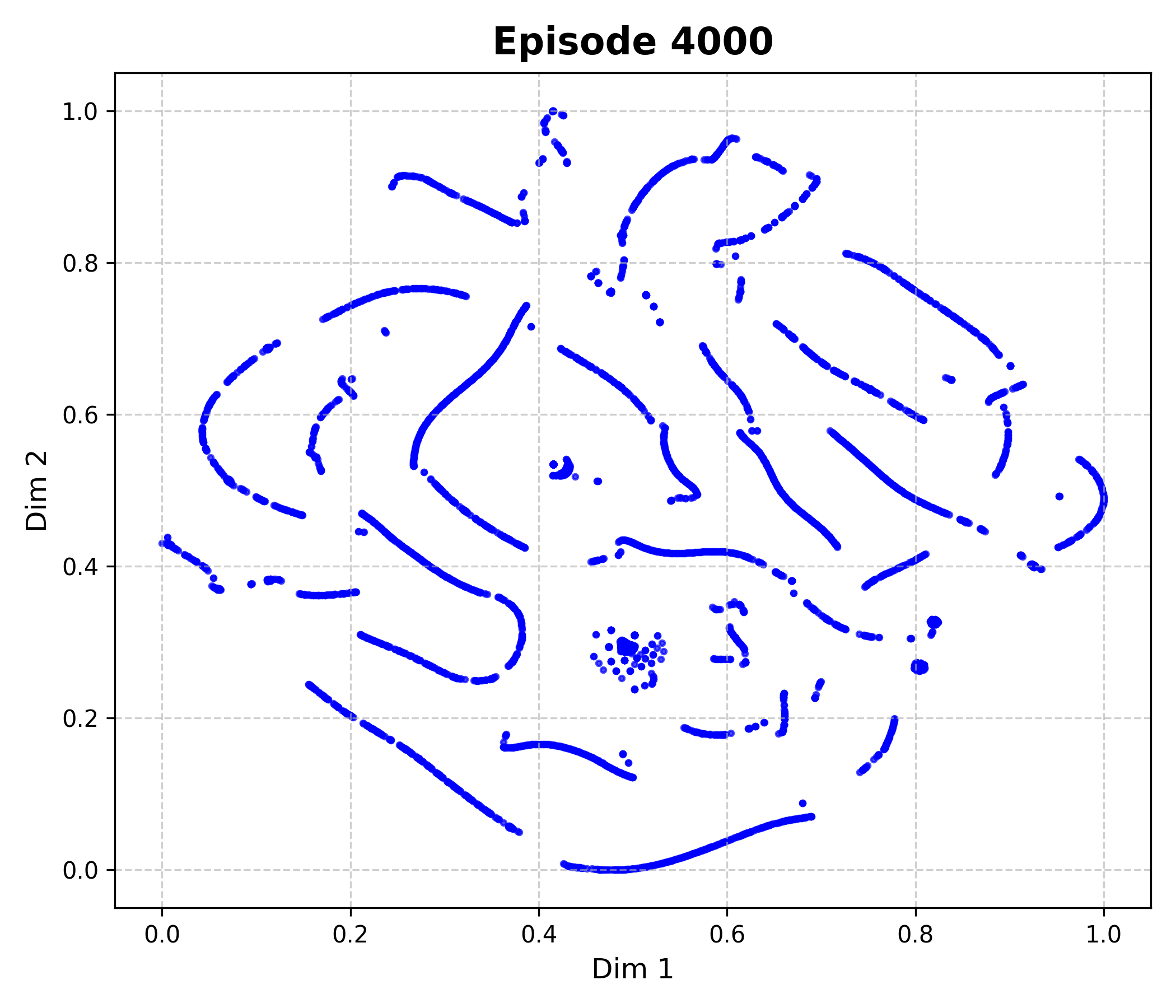}} 
    \subfigure[]{\includegraphics[width=0.3\textwidth]{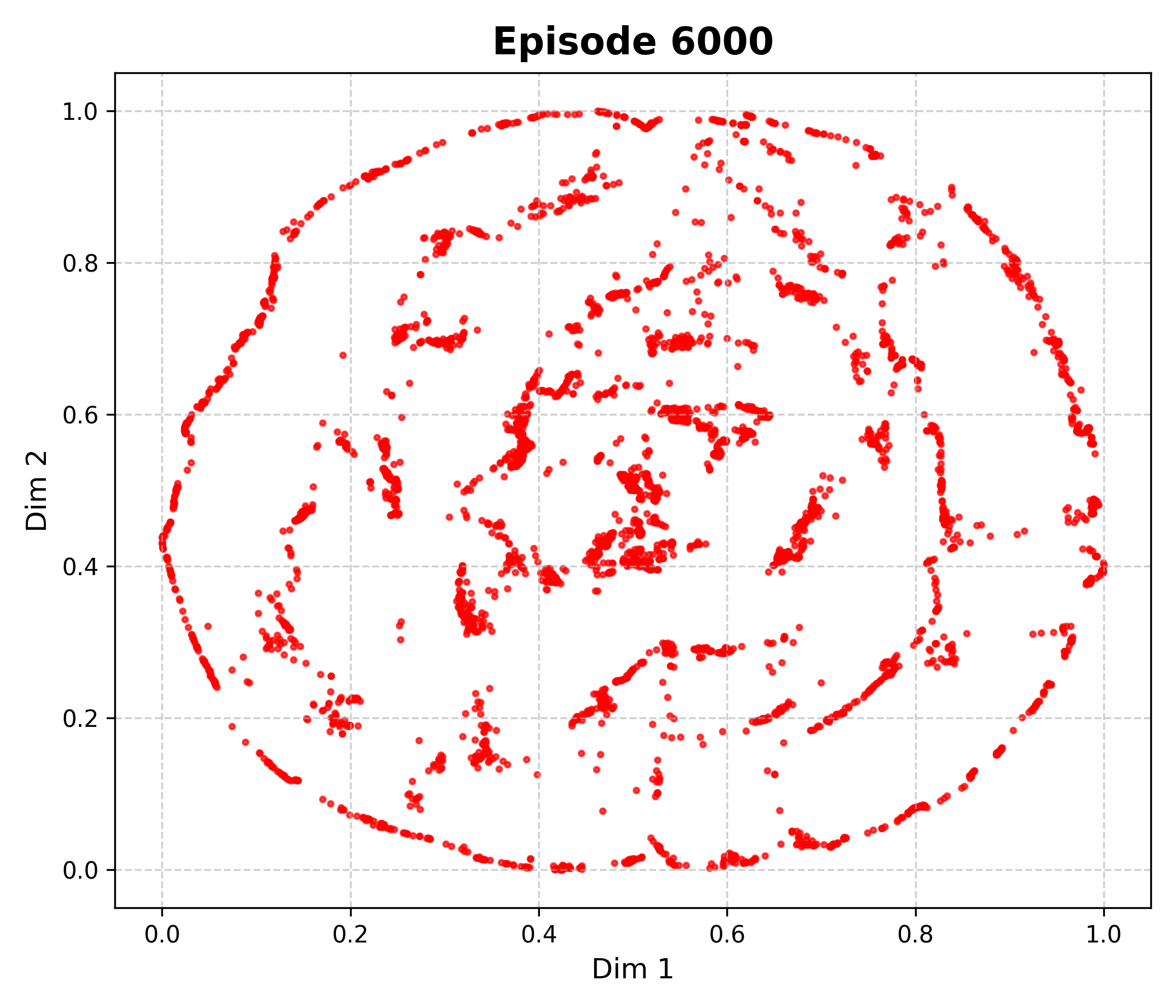}}
    \caption{The distribution of subgoal representations in the subgoal space projected by t-SNE at the (a) 2000th (b) 4000th (c) 6000th episode.}
    \label{fig:subgoal_dist_tsne}
\end{figure}

\begin{figure}[h!]
    \centering
    \subfigure[]{\includegraphics[width=0.3\textwidth]{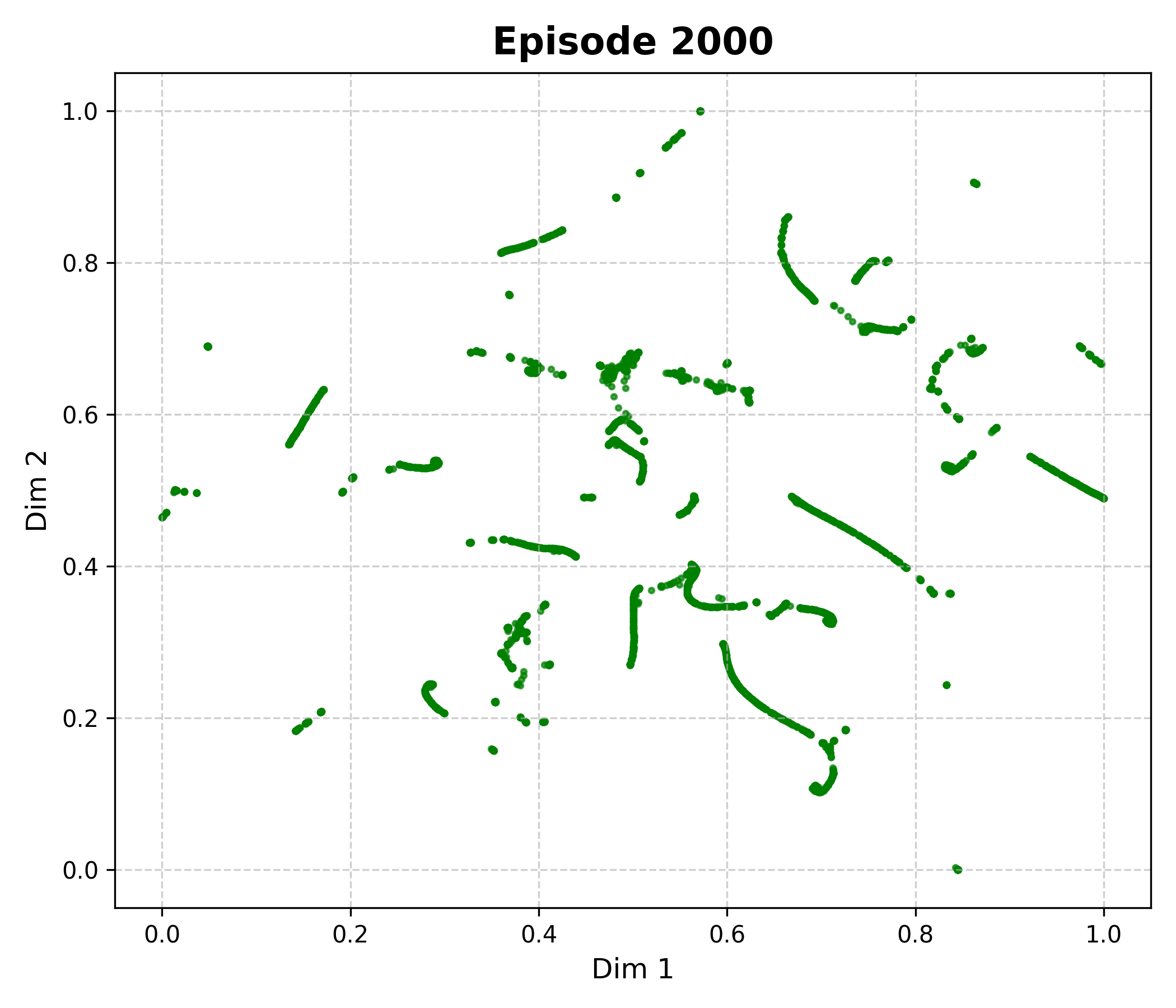}} 
    \subfigure[]{\includegraphics[width=0.3\textwidth]{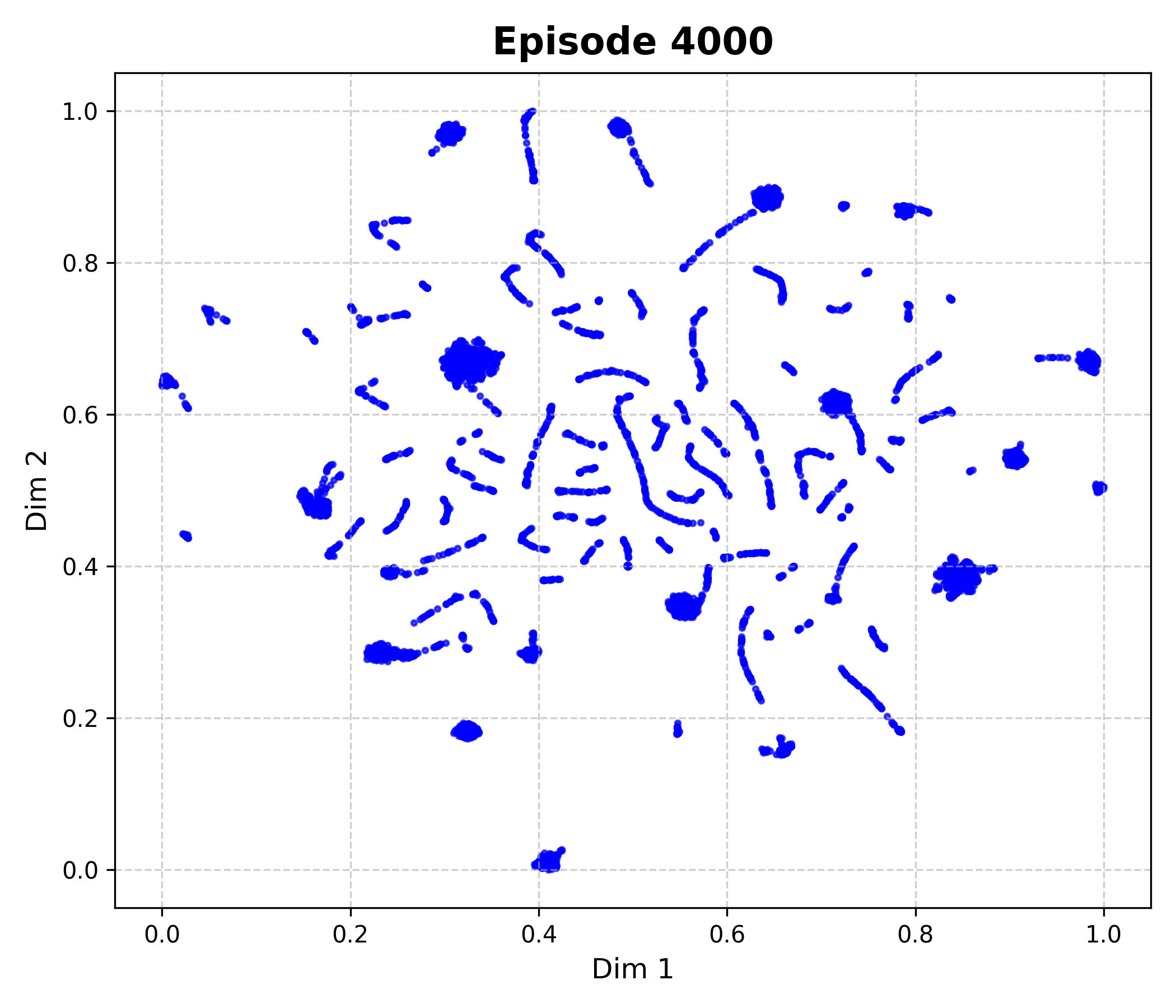}} 
    \subfigure[]{\includegraphics[width=0.3\textwidth]{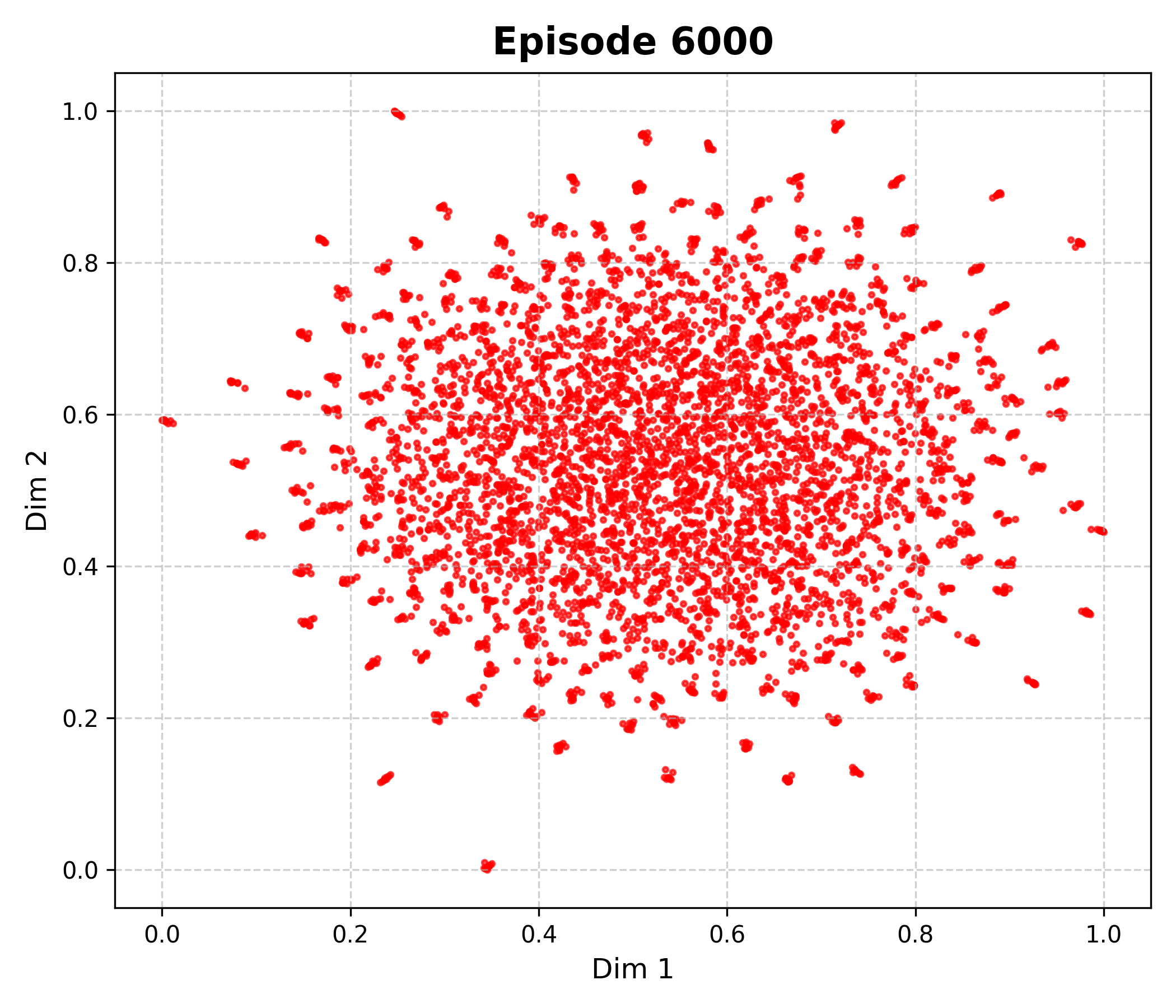}}
    \caption{The distribution of subgoal representations in the subgoal space projected by UMAP at the (a) 2000th (b) 4000th (c) 6000th episode.}
    \label{fig:subgoal_dist_umap}
\end{figure}

In this section, we first extend the visualization results presented in Figure \ref{fig:subgoal_dist} by applying multiple dimensionality reduction techniques—namely PCA, t-SNE, and UMAP—to demonstrate that the observations discussed in Section \ref{sec:vis} are not specific to any single method. The results are shown in Figure \ref{fig:subgoal_dist_pca} to \ref{fig:subgoal_dist_umap}.

While the precise projections vary across methods, a consistent trend emerges: the representations progressively evolve from forming scattered clusters to a more distributed pattern. This trend aligns with the objective of G4RL—to learn a subgoal representation extractor that structures subgoals within a space functioning as a cognitive map, effectively capturing the underlying structure of the original state space. Such a map can be interpreted as an internal model of the environment, offering valuable information for planning and decision-making.

Figures \ref{fig:subgoal_dist_pca} to \ref{fig:subgoal_dist_umap} illustrate how the graph encoder progressively learns to utilize the subgoal space more effectively. Initially, it relies on only a limited region of the space, but over time, its usage becomes more distributed. This progression further supports the claim that the learned subgoal representations are meaningful and hold potential for providing continuous guidance.

\begin{figure}[h!]
    \centering
    \subfigure[]{\includegraphics[width=0.3\textwidth]{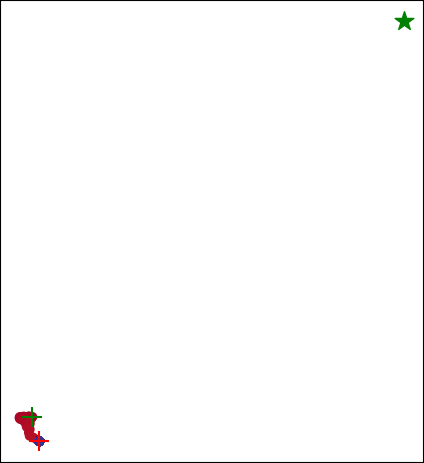}} 
    \subfigure[]{\includegraphics[width=0.397\textwidth]{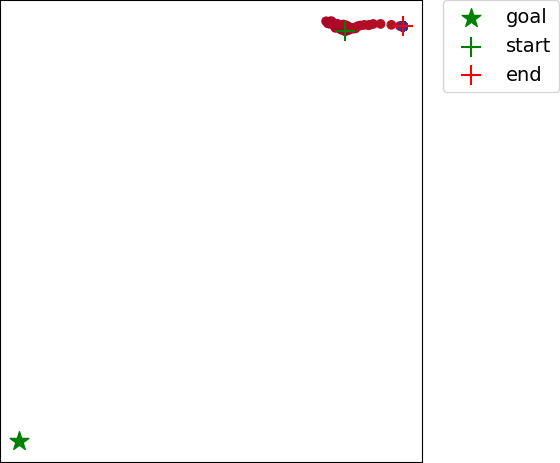}} 
    \caption{Projection of (a) state representations $\phi(s_t)$ in the state space and (b) subgoal representations $g(s_t)$ in the subgoal space at the 2000th episode. The colour spectrum indicates the relative time step (red corresponds to earlier steps), while the green and red crosses mark the projected start and end points, respectively. The green star denotes the projected ultimate goal in the same space.}
    \label{fig:subgoal_dist_2000}
\end{figure}

\begin{figure}[h!]
    \centering
    \subfigure[]{\includegraphics[width=0.3\textwidth]{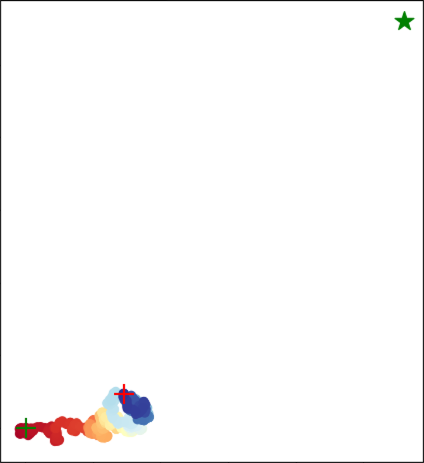}} 
    \subfigure[]{\includegraphics[width=0.4\textwidth]{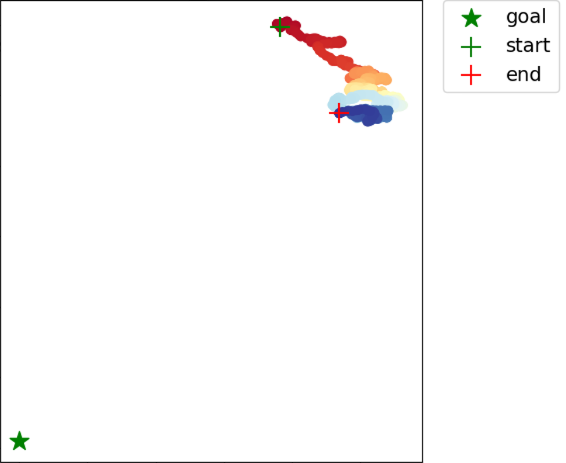}} 
    \caption{Projection of (a) state representations $\phi(s_t)$ in the state space and (b) subgoal representations $g(s_t)$ in the subgoal space at the 4000th episode. The colour spectrum indicates the relative time step (red corresponds to earlier steps), while the green and red crosses mark the projected start and end points, respectively. The green star denotes the projected ultimate goal in the same space.}
    \label{fig:subgoal_dist_4000}
\end{figure}

\begin{figure}[h!]
    \centering
    \subfigure[]{\includegraphics[width=0.3\textwidth]{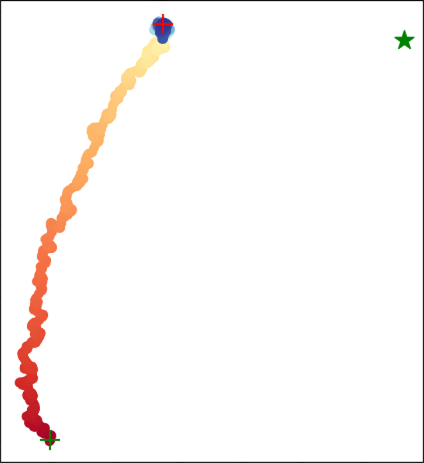}} 
    \subfigure[]{\includegraphics[width=0.399\textwidth]{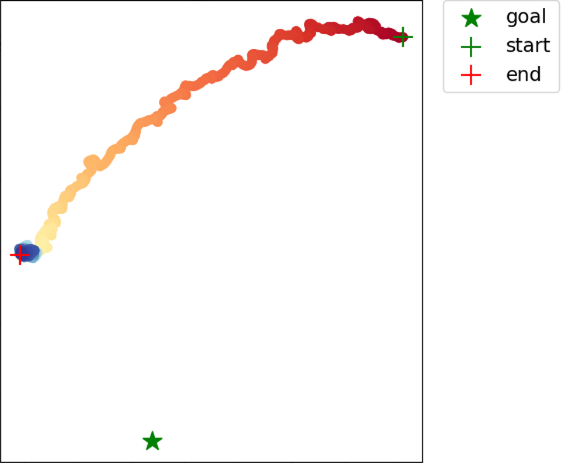}} 
    \caption{Projection of (a) state representations $\phi(s_t)$ in the state space and (b) subgoal representations $g(s_t)$ in the subgoal space at the 6000th episode. The colour spectrum indicates the relative time step (red corresponds to earlier steps), while the green and red crosses mark the projected start and end points, respectively. The green star denotes the projected ultimate goal in the same space.}
    \label{fig:subgoal_dist_6000}
\end{figure}

\begin{figure}[h!]
    \centering
    \subfigure[]{\includegraphics[width=0.3\textwidth]{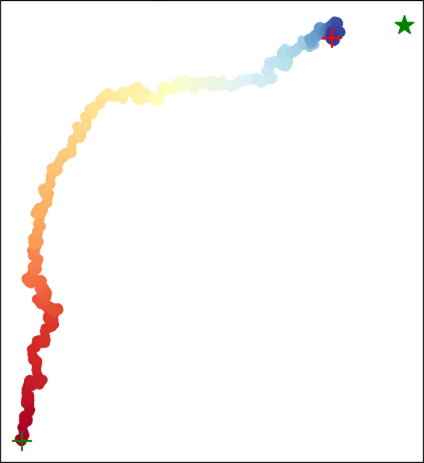}} 
    \subfigure[]{\includegraphics[width=0.398\textwidth]{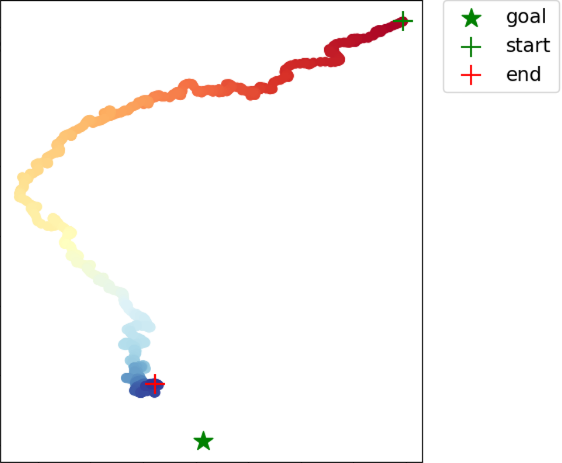}} 
    \caption{Projection of (a) state representations $\phi(s_t)$ in the state space and (b) subgoal representations $g(s_t)$ in the subgoal space at the 8000th episode. The colour spectrum indicates the relative time step (red corresponds to earlier steps), while the green and red crosses mark the projected start and end points, respectively. The green star denotes the projected ultimate goal in the same space.}
    \label{fig:subgoal_dist_8000}
\end{figure}

We then visualize trajectories sampled at different stages of training in the AntMaze environment, projecting $\phi(s_t)$ into the state space and $g(s_t)$ into the subgoal space, respectively. Results are shown in Figures \ref{fig:subgoal_dist_2000} to \ref{fig:subgoal_dist_8000}.


Figures \ref{fig:subgoal_dist_2000} to \ref{fig:subgoal_dist_8000} demonstrate that the learned subgoal space evolves progressively to reflect the agent’s observations of the state space. A clear correspondence emerges between the connection structure of state representations $\phi(s_t)$ in the state space and subgoal representations $g(s_t)$ in the subgoal space. This supports the claim that the subgoal space captures the connectivity of the original state space and that the proposed subgoals offer accurate and consistent guidance toward the final goal.
\end{document}